\documentclass{article}

\usepackage{microtype}
\usepackage{graphicx}
\usepackage{subfigure}
\usepackage{booktabs} 
\usepackage{array}
\usepackage{hyperref}

\usepackage[accepted]{icml2024}

\usepackage{amsmath}
\usepackage{amssymb}
\usepackage{mathtools}
\usepackage{amsthm}
\usepackage{threeparttable}
\usepackage{tabularx}

\usepackage[capitalize,noabbrev]{cleveref}

\theoremstyle{plain}

\theoremstyle{definition}

\theoremstyle{remark}

\usepackage[textsize=tiny]{todonotes}

\icmltitlerunning{LoRAP: Transformer Sub-Layers Deserve Differentiated Structured  Compression for  Large Language Models}
\usepackage{multirow}
\begin{document}

\twocolumn[
\icmltitle{LoRAP: Transformer Sub-Layers Deserve Differentiated Structured  Compression for  Large Language Models}

\icmlsetsymbol{equal}{*}
\author{Author One}
\centering
\textbf{Guangyan Li}\textsuperscript{1}  \textbf{,Yongqiang Tang}\textsuperscript{1}  \textbf{,Wensheng Zhang}\textsuperscript{1}\\
\textsuperscript{1} Institute of Automation, Chinese Academy of Sciences,
Beijing 100190, China \\
\{liguangyan2022,yongqiang.tang\}@ia.ac.cn \\
zhangwenshengia@hotmail.com




\icmlkeywords{Machine Learning, ICML}

\vskip 0.3in
]





\begin{abstract}
Large language models (LLMs) show excellent performance in difficult tasks, but they often require massive memories and computational resources. How to reduce the parameter scale of LLMs has become research hotspots. In this study, we make an important observation that the multi-head self-attention (MHA) sub-layer of Transformer exhibits  noticeable low-rank structure,  while the feed-forward network (FFN) sub-layer does not. 
With this regard, we design a mixed  compression model, which organically combines \textbf{L}\textbf{o}w-\textbf{R}ank matrix approximation \textbf{A}nd structured \textbf{P}runing (\textbf{LoRAP}). For the MHA sub-layer,  we propose an input activation weighted singular value decomposition method to strengthen the low-rank characteristic. 
Furthermore, we discover that the weight matrices in MHA sub-layer have different low-rank degrees. Thus, a novel parameter allocation scheme according to the discrepancy of low-rank degrees is devised. For the FFN sub-layer, we propose a gradient-free structured channel pruning method.  During the pruning, we get an interesting finding that the least important 1\% of parameter actually play a vital role in model performance. Extensive  evaluations on zero-shot perplexity and zero-shot task classification indicate that our proposal  is superior to previous structured compression rivals under multiple compression ratios. 
\end{abstract}

\section{Introduction}
\label{submission}
Large language models (LLMs) \cite{GLM,LLaMA,Vicuna} have revolutionized the field of natural language processing (NLP), exhibiting significant advancements in language understanding and generation \cite{brown2020language}. As the size increases, LLMs can handle more complex tasks and even display emergent abilities \cite{emergent_ability}. However, the large size of these models poses challenges in deployment and inference, which require massive memory and computational resources. For instance, the largest LLaMA \cite{LLaMA} consists of  70 billion parameters and ChatGPT is even larger at the scale of 175 billion parameters.

There have been plentiful techniques to compress the transformer-based  models, including pruning \cite{magitude-prune,Cofi,OBC}, low-rank approximation \cite{first-SVD,FWSVD-2}, quantization \cite{GPTQ,zeroquant,QLORA}, and knowledge distillation \cite{distill,SCOTT}. The traditional compression methods usually require fine-tuning the compressed model on concrete tasks to recovery the specific capability of the model. Nevertheless,  the compression for LLMs primarily concentrates on reducing the model size while retaining the general capability \cite{LLM-pruner}. Therefore, rudely applying previous transformer-based compression methods without specialized consideration might compromise their capacities for generic tasks \cite{zero-shot-learners}. More recently, by virtue of the advantage of directly reducing the  parameter scale, low-rank approximation and pruning  have attracted much attention in the area of large model compression.

Low-rank approximation reduces the  parameter size via decomposing the original weight matrix into two smaller matrices.
In \cite{first-SVD}, the authors perform Singular Value Decomposition (SVD) on BERT and utilize knowledge distillation to recovery model performance. DRONE \cite{DRONE} approximates input activations through SVD  in specific tasks. 
FWSVD \cite{FWSVD} evaluates the importance  of weights with the Fisher information and conducts SVD on the weighted  matrix.
AFM \cite{AFM} decomposes the weight matrix based on the low-rank property of the output activations.
LORD \cite{LORD}  compresses a 16B code model with AFM and then restores its performance  by LoRA fine-tuning. Very recently, several efforts attempt to incorporate low-rank approximation with other compression techniques. LoftQ \cite{LoftQ} utilizes a quantized matrix and two low-rank matrices to approximate the original high-precision weight matrix. In \cite{Losparse}, the  weight matrix is approximated by the sum of a low-rank matrix and a sparse matrix. LPAF \cite{prune-deco} performs SVD on the pruned model obtained through movement pruning. Despite the brilliant achievements, the present works typically  compress each module of Transformer layer in the same manner, while ignoring a fundamental problem, that is, \emph{whether the modules in transformer have the same property}.  This beneficial exploration is expected to provide important guidance for the improvement of compression methods.

Pruning aims to remove unimportant parts of the weights, which can be categorized into unstructured pruning and structured pruning. Unstructured pruning entails the removal of less important individual weights based on their importance scores. The representative unstructured pruning methods for LLMs include SparseGPT \cite{sparsegpt}, Wanda \cite{wanda}, GBLM-pruner \cite{GBLM-pruner}. Although unstructured pruning yields favorable performance, the acceleration in inference is only achievable on specific hardware due to the irregular sparsity, which makes it difficult to migrate across different platforms and environments. In contrast, structured pruning eliminates the weights according to a certain structure or pattern, such as channel, attention head or layer. This enables it to reduce storage memory and accelerate inference on common hardware.
The existing structured pruning research usually estimates the importance score based on the gradient. For example, LLM-Pruner \cite{LLM-pruner} adopts the Taylor expansion of loss function to measure the importance. 
LoRAPrune \cite{LORApruner} approximates its weight gradients with LoRA \cite{LORA} weights during the LoRA fine-tuning process. 
LoRAShear \cite{LORAshear} establishes dependency graph for grouping weights and then adopts progressive pruning strategy on LoRA adaptors. 
These gradient-based methods require either substantial storage and computation resources or intricate pruning steps. In the structured pruning, \emph{how to achieve  gradient-free and meaningful importance estimation for weights}  becomes a valuable but less studied issue.

\textbf{Contributions.} Our contributions are as follows:
\begin{itemize}
\item We analyze the distribution of important weights in different Transformer sub-layers of the LLaMA  model and observe that, the multi-head self-attention (MHA) sub-layer exhibits a more pronounced low-rank property than feed-forward network (FFN) sub-layer.  This inspires us to combine \textbf{Lo}w-\textbf{R}ank matrix approximation \textbf{A}nd structure \textbf{P}runing (\textbf{LoRAP}), with each  compressing the MHA and FFN sub-layers separately.

\item For the MHA sub-layer, we propose an Activation Weighted SVD (AWSVD) method, which evaluates the weight importance  in terms of the $\ell_2$ norm of the corresponding input activations. Besides, we find that the weight matrices in MHA sub-layer have varying low-rank degrees. Thus, we  propose to allocate more parameters to weight matrices with poorer low-rank property.

\item For the FFN sub-layer, we devise a gradient-free structured pruning method, which removes the associated channels according to the group importance. During the channel pruning process, we discover that the least important parameters (approximately 1\%) surprisingly play a crucial role in the model's performance. Therefore, we suggest to retain these parameters under a fixed parameter budget.
\item 
We evaluate the performance of the compressed model through zero-shot perplexity on WikiText2 and PTB datasets, as well as zero-shot task classification on 7 common-sense reasoning datasets. At multiple compression ratios, our method outperforms existing structured pruning and low-rank approximation methods. 
\end{itemize}



\section{Background}
We briefly review the transformer architecture, the importance scores of weights, structured pruning and low-rank approximation.

\subsection{Transformer Model}
Models based on the transformer architecture usually consist of several consecutive layers, with each layer comprising a multi-head self-attention  (MHA) sub-layer and a fully connected feed-forward network (FFN) sub-layer. We use the LLaMA model as an example to introduce the transformer. Given the input activation $\mathbf{X} \in \mathbb{R}^{L \times d}$ obtained by the layer normalization, where $L$ and $d$ are sequence and feature dimensions respectively, the forward computation for MHA is as follows:
\begin{align*}
  & \mathrm{head}_{i}= \mathrm{Softmax}(\mathbf{X}\mathbf{W}_{q_{i}}(\mathbf{X}\mathbf{W}_{k_{i}})^{\mathrm{T}}/\sqrt{d_{h}})\mathbf{X}\mathbf{W}_{v_{i}} \\
  & \mathrm{MHA}(\mathbf{X})= \mathrm{Concat}( \mathrm{head}_{1}, \dots, \mathrm{head}_{h})\mathbf{W}_{o},
\end{align*}
where $\mathbf{W}_{q_{i}}$, $\mathbf{W}_{k_{i}}$, $\mathbf{W}_{v_{i}} \in \mathbb{R}^{d\times d_{h}} $ are query, key, and value matrices in $i$-th head, 
$\mathbf{W}_{o} \in \mathbb{R}^{d \times d}$ denotes an output projection matrix. $h$ and $d_{h}$ represent the number of heads and the dimension of each head in multi-head attention, respectively.  where $\mathbf{W}_{q_{i}}$, $\mathbf{W}_{k_{i}}$, $\mathbf{W}_{v_{i}} \in \mathbb{R}^{d\times d_{h}} $ are query, key, and value matrices in $i$-th head,  $\mathbf{W}_{o} \in \mathbb{R}^{d \times d}$ denotes an output projection matrix. $h$ and $d_{h}$ represent the number of heads and the dimension of each head in multi-head attention, respectively. In general, we have $d = d_h \times h$.
The output of the MHA sub-layer serves as the input of the FFN sub-layer. 
FFN sub-layer comprises three linear transformations and an activation, the forward computation is as follows:
\begin{align*}
  & \mathrm{FFN}(\mathbf{X}) = (\mathbf{X}\mathbf{W}_{up} \odot \sigma(\mathbf{X}\mathbf{W}_{gate}))\mathbf{W}_{down},
\end{align*}
where $\mathbf{W}_{up}, \mathbf{W}_{gate} \in \mathbb{R}^{d \times d_{m}}$, $\mathbf{W}_{down} \in \mathbb{R}^{d_{m} \times d} $ and $\sigma(
\cdot)$ is Silu activation function. And $\odot$ represent the Hadamard product. In the subsequent context, without loss of generality, we define all the matrix multiplication as $\mathbf{y} = \mathbf{W} \mathbf{x}$, where 
 $\mathbf{W} \in \mathbb{R}^{d_{out}\times d_{in}}$  denote the weight matrix in the model. 

\subsection{Importance Score of Weights}
It is observed that in LLMs with model size of 6B or more, a distinct subset of activations exhibit significantly larger magnitudes compared to the remaining activations. These specific activations  are critical to the performance of the model and are denoted as outlier activations \cite{LLM-int8}. The magnitude of the activation value can reflect the importance score of the weights.
Wanda \cite{wanda} is the first method to estimate the importance score of weights by 
activations in   pruning. Specifically, they calculate the product of the weight magnitude and the corresponding input activation's $\ell_2$ norm as the importance score. The calculation formulation is as follows:
\begin{equation}\label{weight-importance}
   I(W_{ij}) = |W_{ij}| \cdot \| \mathbf{X}_{j}\|_{2},  
\end{equation}
where $I(W_{ij})$ is the importance score function of weight $W_{ij}$, $|\cdot|$ is the absolute value operator, $W_{ij}$ represents the $ij$-th entry of $\mathbf{W}$.  $\mathbf{X}_{j} \in \mathbb{R}^{N\times L}$ denotes the $j$-th feature aggregated across $N$ input samples.
It's worth noting that the calculation is gradient-free, thereby mitigating the demand for memory and computational resources.

\subsection{Structured Pruning}
The structured pruning targets different pruning granularities, including layer pruning, attention head pruning, and channel pruning. Previous pruning methods \cite{MVP,GRAIN} estimate importance scores based on gradients, which needs massive 
 computational resources and makes them challenging to be applied to LLMs. The importance scores in Eq. (\ref{weight-importance}) is gradient-free, which avoids the calculation and storage of gradients. However, it is an importance score at the level of individual weights, which cannot be directly applied to structured pruning. LoSparse \cite{Losparse} introduces neuron importance scores into structured pruning. The importance score function $\Phi(\mathbf{W}_{i,:})$  of the $i$-th output neuron is calculated as follows:
\begin{align}\label{channel-importance-old}
    \Phi(\mathbf{W}_{i,:})=\frac{1}{d_{in}} \sum\limits_{j = 1}^{d_{in}} I(W_{ij}). 
\end{align}
Note that beyond Eq.(\ref{channel-importance-old}) that adopts the arithmetic mean as the score function, various other formulations can also be employed, such as $\ell_2$ norm, $\ell_{\infty}$ norm.

\begin{figure}[h]
  \centering
  \includegraphics[width=0.48\textwidth]{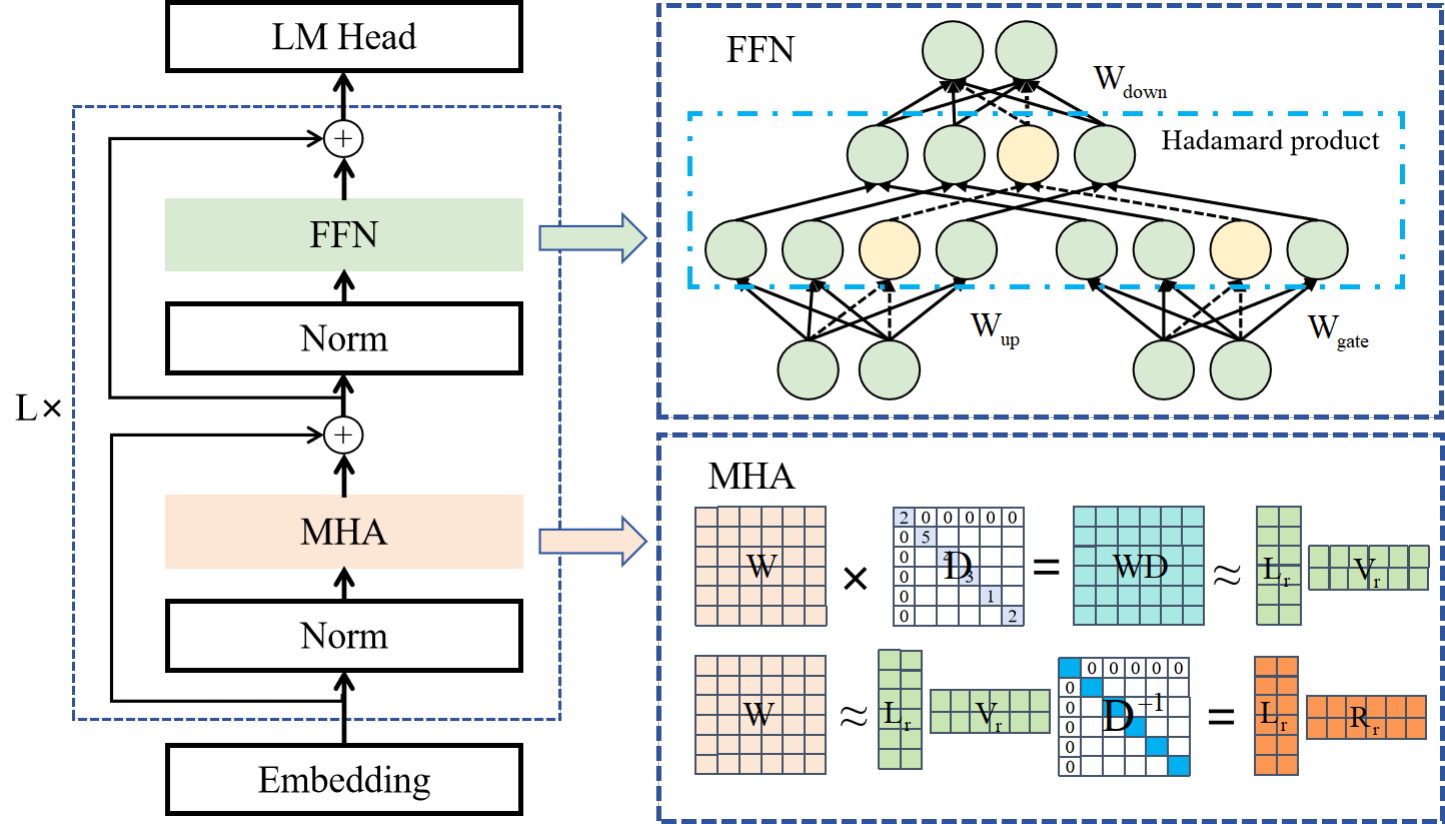}
  \vskip -0.07in
  \caption{The compression of the transformer layer. For the FFN sub-layer, we prune the neurons in the intermediate layer. For the MHA sub-layer, we employ weighted SVD to obtain two low-rank matrices  as an approximation to the original matrix.}
  \label{main struct}
  \vskip -0.1in
\end{figure}

\subsection{Low-Rank Approximation of Matrix}
Low-rank approximation substitutes the original weight matrix $\mathbf{W}$ with two lower-rank matrices $\mathbf{L} \in \mathbb{R}^{d_{out} \times r}$ and $\mathbf{R} \in \mathbb{R}^{r \times d_{in}}$. 
 Given the input $\mathbf{x}$, the output $\mathbf{y}$ is 
\begin{equation}
    \mathbf{y}=\mathbf{Wx} \approx \mathbf{LRx}.
\end{equation}
SVD is the most commonly used matrix factorization method, which offers the best $r$-rank approximation of the matrix. The initial matrix $\mathbf{W} \in \mathbb{R}^{d_{out} \times d_{in}}$ is  decomposed into $\mathrm{SVD}(\mathbf{W})=\mathbf{U}\mathbf{\Sigma}\mathbf{V}$, where $\mathbf{U} \in \mathbb{R}^{d_{out} \times d_{out}}$ and $\mathbf{V} \in \mathbb{R}^{d_{in} \times d_{in}}$ are orthogonal matrix, and  $\mathbf{\Sigma} \in \mathbb{R}^{d_{out} \times d_{in}}$  is a matrix with only non-zero values on the diagonal. By retaining the largest $r$ eigenvalues, we can decompose $\mathbf{W}$ into two low-rank matrices $\mathbf{L}$ and $\mathbf{R}$ as follows:
\begin{equation}
   \mathbf{W} \approx (\mathbf{U}_{r}\mathbf{\Sigma}_{r})\mathbf{V}_{r}=\mathbf{LR}.
\end{equation}

\section{The Proposed Method}
We propose a novel method for compressing LLMs. Specifically, we compress the different sub-layers in model with low-rank approximation and structured pruning separately. The method is illustrated in Fig. \ref{main struct}. 

\subsection{Weight Distribution in MHA and FFN Sub-Layers}

\begin{figure}[h]
  \centering
  \subfigure[$\mathbf{W}_{q}$]{\includegraphics[width=0.2\textwidth]{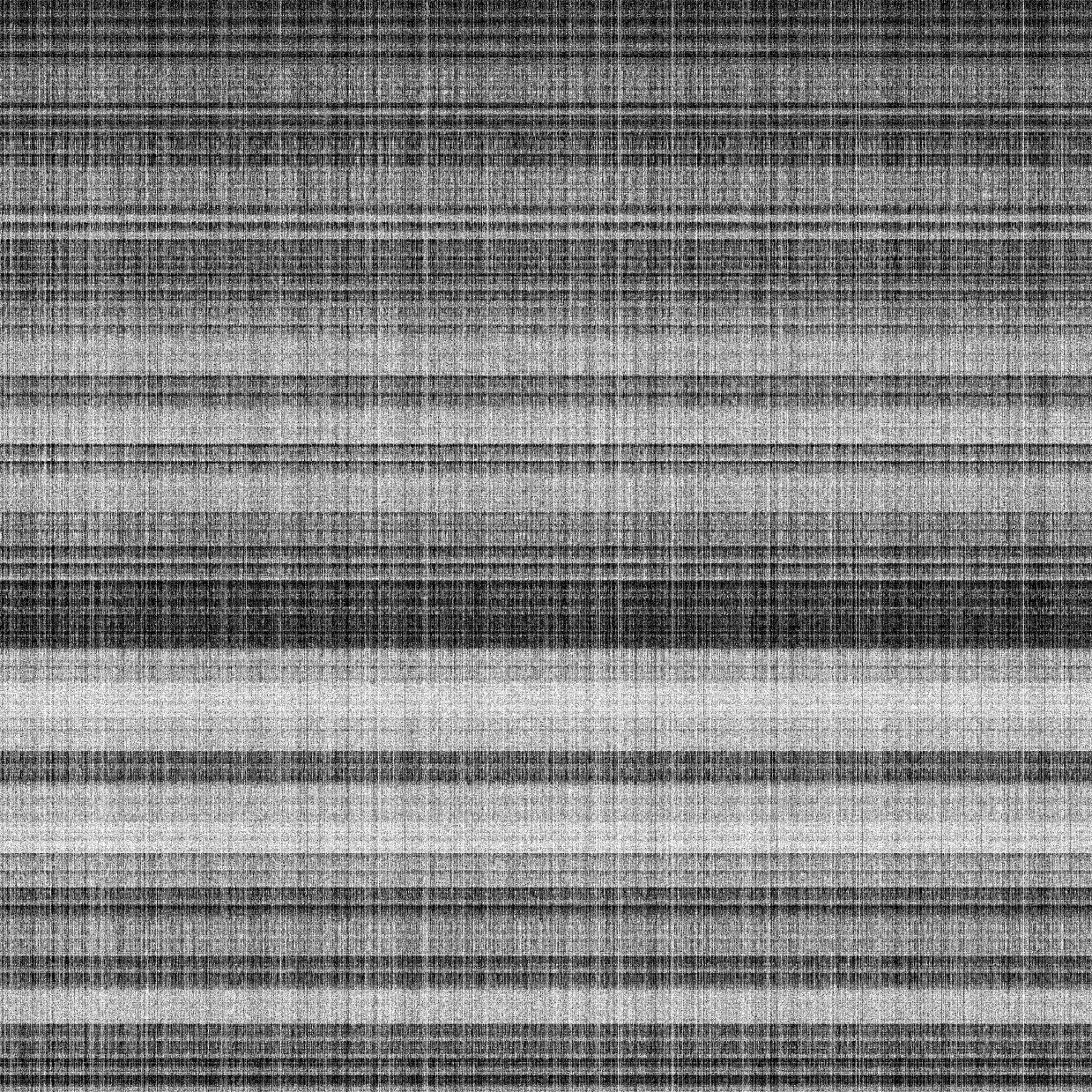}}
  \vspace{-5pt}  
  \hspace{5pt} 
  \subfigure[$\mathbf{W}_{k}$]{\includegraphics[width=0.2\textwidth]{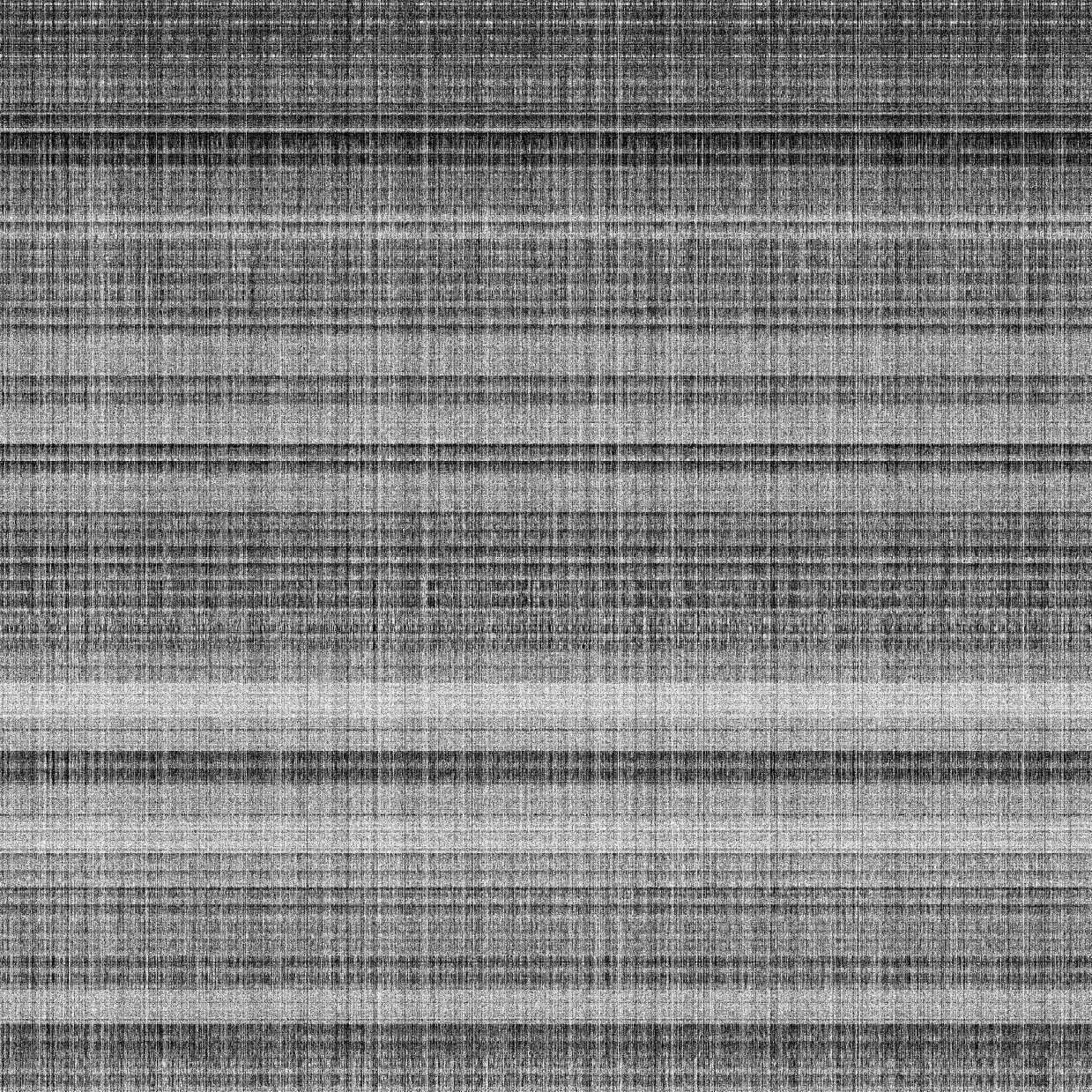}}
  \vspace{-10pt}  
  \subfigure[$\mathbf{W}_{v}$]{\includegraphics[width=0.2\textwidth]{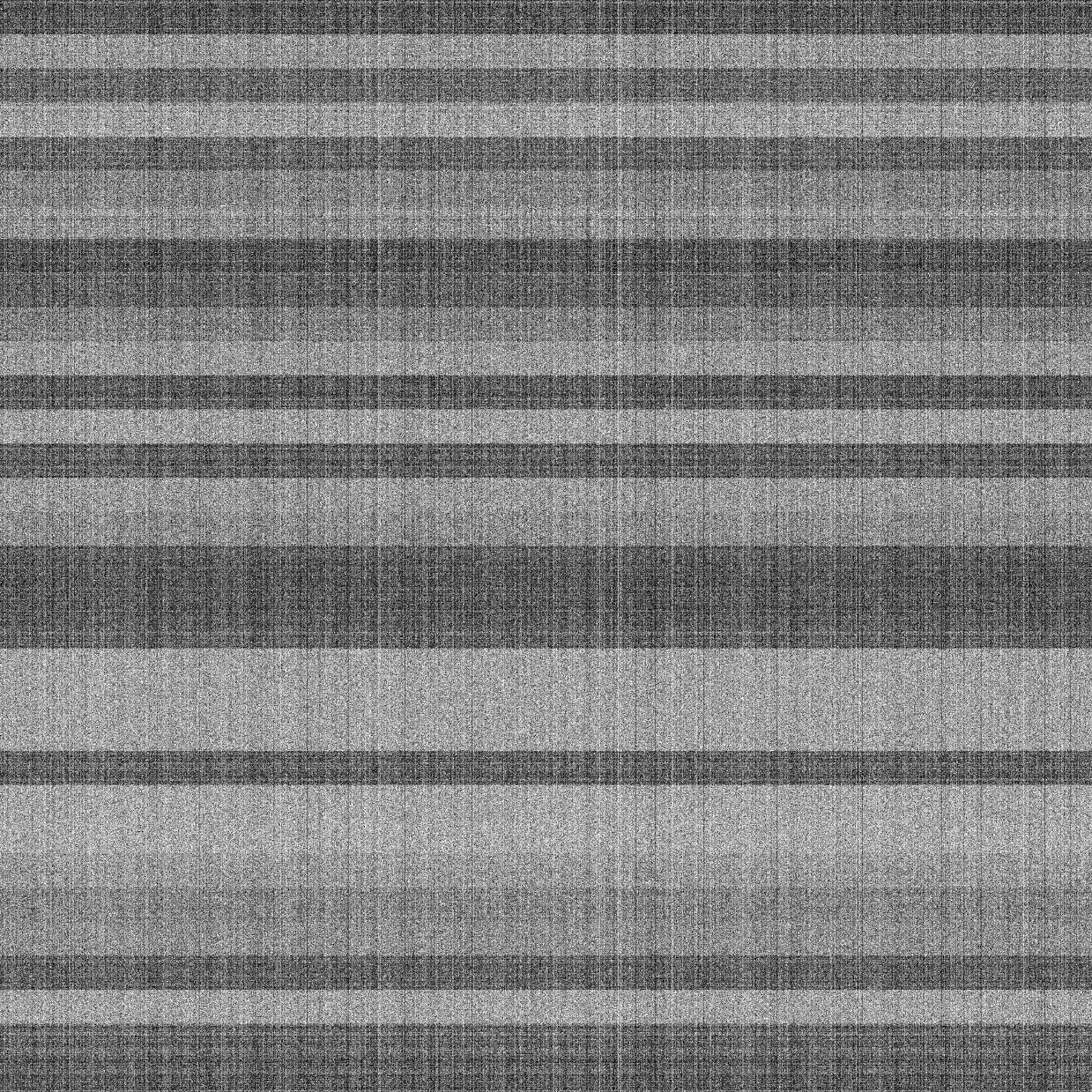}}
  \hspace{5pt} 
  \subfigure[$\mathbf{W}_{o}$]{\includegraphics[width=0.2\textwidth]{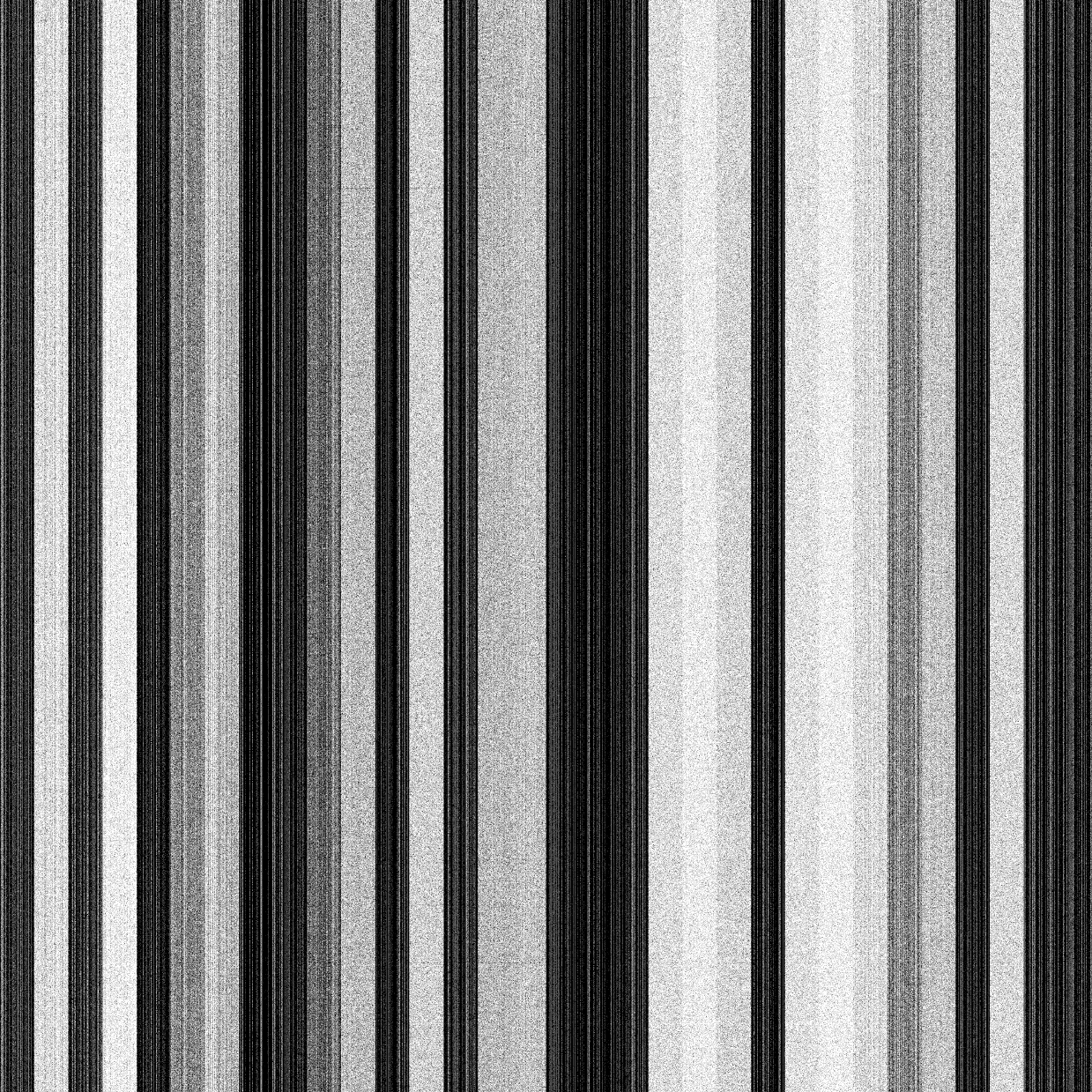}}
  \caption{Visualization of the $\mathbf{W}_{q}$, $\mathbf{W}_{k}$, $\mathbf{W}_{v}$, and $\mathbf{W}_{o}$ matrices in the first MHA sub-layer at 50\% sparsity. The black areas in the figure represent the pruned weights, while the white areas indicate the retaining weights.}
  \label{wanda visualization}
\end{figure}

As we know, unstructured pruning has poor generality but strong performance, whereas structured pruning has strong generality but typically slightly worse performance.
In this study, we aim to design a structured compression method that could achieve promising performance closer to unstructured pruning, meanwhile ensure its generalization on common devices. The weights obtained from unstructured pruning represent the crucial weights that impact the performance of model. Analyzing the structural patterns therein can better guide structured compression.

\begin{figure}[h]
  \centering
  {\includegraphics[width=0.48\textwidth]{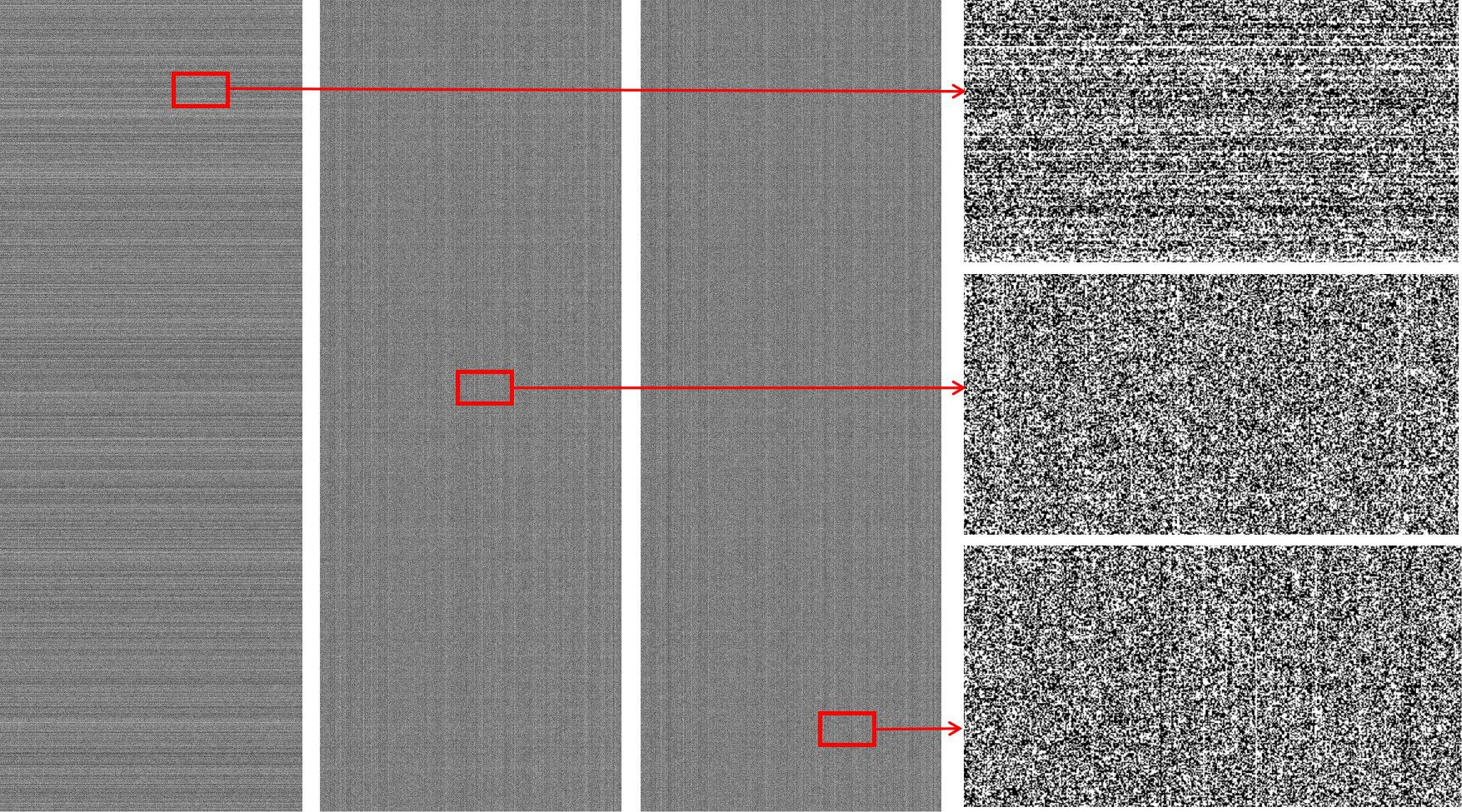}}
  \vskip -0.05in
  \caption{The three images on the left are visualization of the  $\mathbf{W}_{down}$, $\mathbf{W}_{gate}$, and $\mathbf{W}_{up}$ matrices (From left to right). The three images on the right are areas of size $800\times420$.}
  \label{FFN visualition}
\end{figure}

After unstructured pruning with Wanda under 50\% sparsity,  the weight matrices of the first MHA sub-layer are visualized in Fig. \ref{wanda visualization}. We observe an interesting phenomenon, that is, 
the distribution of retained weights is more concentrated in some certain rows or columns, 
\begin{table}[htb]
\caption{The ratio of singular values when 80\% energy is retained.}
  \centering
  \setlength\tabcolsep{5pt}
  \scriptsize
  \vskip 0.1in
  \begin{threeparttable}
  \begin{tabular}{cccccccc}
    \toprule
    Method & $\mathbf{W}_{q}$ & $\mathbf{W}_{k}$  & $\mathbf{W}_{v}$ & $\mathbf{W}_{o}$ & $\mathbf{W}_{gate}$ & $\mathbf{W}_{up}$ & $\mathbf{W}_{down}$ \\
    \midrule
        Original  & 16.60 & 15.36	& 43.90 & 45.53	& 53.42 & 54.49	& 63.72 \\
        AWSVD  & 5.86 & 5.15 & 42.19 & 29.52 & 52.12 & 53.52 & 59.72 \\
    \bottomrule
  \end{tabular}
 \vskip -0.1in
  \label{low-rank-explore}
  \end{threeparttable}
\end{table}
We further present the visualization of the FFN sub-layer in Figure \ref{FFN visualition}. Intuitively, the weight distribution of FFN is different from MHA. More strictly speaking, FFN exhibits less low-rank characteristic. To quantitatively analyze the low-rank property, we perform SVD on the weight matrices of MHA and FFN sub-layers. The first row in Table \ref{low-rank-explore} shows the percentage of the number of singular values when 80\% energy is retained. The results demonstrate that compared to FFN sub-layer, the weights in MHA sub-layer exhibit a more pronounced low-rank structure. This suggests that low-rank approximation is more suitable to the MHA sub-layer and the FFN sub-layer deserves other compression strategies. Based on this insight, in this paper, we propose a weighted low-rank approximation method to compress the MHA sub-layer and adopt the structured pruning to compress the FFN sub-layer. In the following, we will elaborate our proposal.

\subsection{Weighted Low-Rank Approximation for MHA}
It is observed  that model with weighted SVD can achieve better performance than the original SVD \cite{FWSVD}. We draw inspiration from Wanda to estimate the importance scores of weights with the activation values  and propose a Activation Weighted SVD (AWSVD) method.
For each individual weight $W_{ij}$, we estimate its importance score based on the $\ell_2$ norm of the corresponding input activations (see Eq. (\ref{weight-importance})). We use 
\begin{equation}\label{x_din}
\mathbf{x}_{d_{in}}=(\|\mathbf{X}_{1}\|_2, \|\mathbf{X}_{2}\|_2, \cdots ,\|\mathbf{X}_{d_{in}}\|_{2}) 
\end{equation}
to represent the vector of importance scores, where $\|\mathbf{X}_{j}\|_{2}$ is the importance score of the $j$-th column  of weight matrix $\mathbf{W}_{:,j}$. The  weighted reconstruction error is as follows:
\begin{equation}\label{weighted SVD}
  {\underset{\mathbf{L}, \mathbf{R}} {\mathrm{min}}} \ \ {\underset{i,j}  {\sum}} 
  ({W_{ij}}-(\mathbf{LR})_{ij})^2 \|\mathbf{X}_{j}\|_{2},
\end{equation}
Where $\mathbf{L}$ and $\mathbf{R}$ are obtained by matrix decomposition.
 In formulation (\ref{weighted SVD}), since each column in the matrix $\mathbf{W}$ has equal importance score, with this characteristic, we can directly obtain its closed-form solution. We define the importance scores as the diagonal matrix $\mathbf{D}=\mathrm{diag}(\mathbf{x}_{d_{in}})$. The problem of Eq. (\ref{weighted SVD}) can be rewritten as:
\begin{equation}\label{FWSVD}
  {\underset{\mathbf{L},\mathbf{R}}{\mathrm{min}}} \ \ \|\mathbf{\mathbf{WD}}-\mathbf{LRD}\|_{2}.
\end{equation}
Performing standard SVD on the weighted matrix $\mathbf{WD}$ yields the result: SVD$(\mathbf{WD})=\mathbf{U} \mathbf{\Sigma} \mathbf{V}$. 
Therefore, the solution of Eq. (\ref{FWSVD}) is $\mathbf{L}=\mathbf{U}\mathbf{\Sigma},\mathbf{R}=\mathbf{VD}^{-1}$. 
To compress the weighted matrix, we retain the first $r$ components of the matrices $\mathbf{L}$ and $\mathbf{R}$. In the end, we obtain $\mathbf{L}_{r}=\mathbf{U}_{r}\mathbf{\Sigma}_{r},\mathbf{R}_{r}=\mathbf{V}_{r}\mathbf{D}^{-1}$.
At the second row in Table \ref{low-rank-explore}, we show the percentage of the number of singular values obtained by our AWSVD when 80\% energy is retained. We observe that  the weighted matrix exhibits a stronger low-rank property when compared with the original matrix, which implies that weighted SVD may achieve higher compression ratios.
Besides, compared to FFN, the improvement in MHA are more significant. This further confirms that the low-rank approximation is more suitable for MHA sub-layer.

\begin{figure}[h]
  \centering
  \subfigure[PPL on WikiText2]{\includegraphics[width=0.23\textwidth]{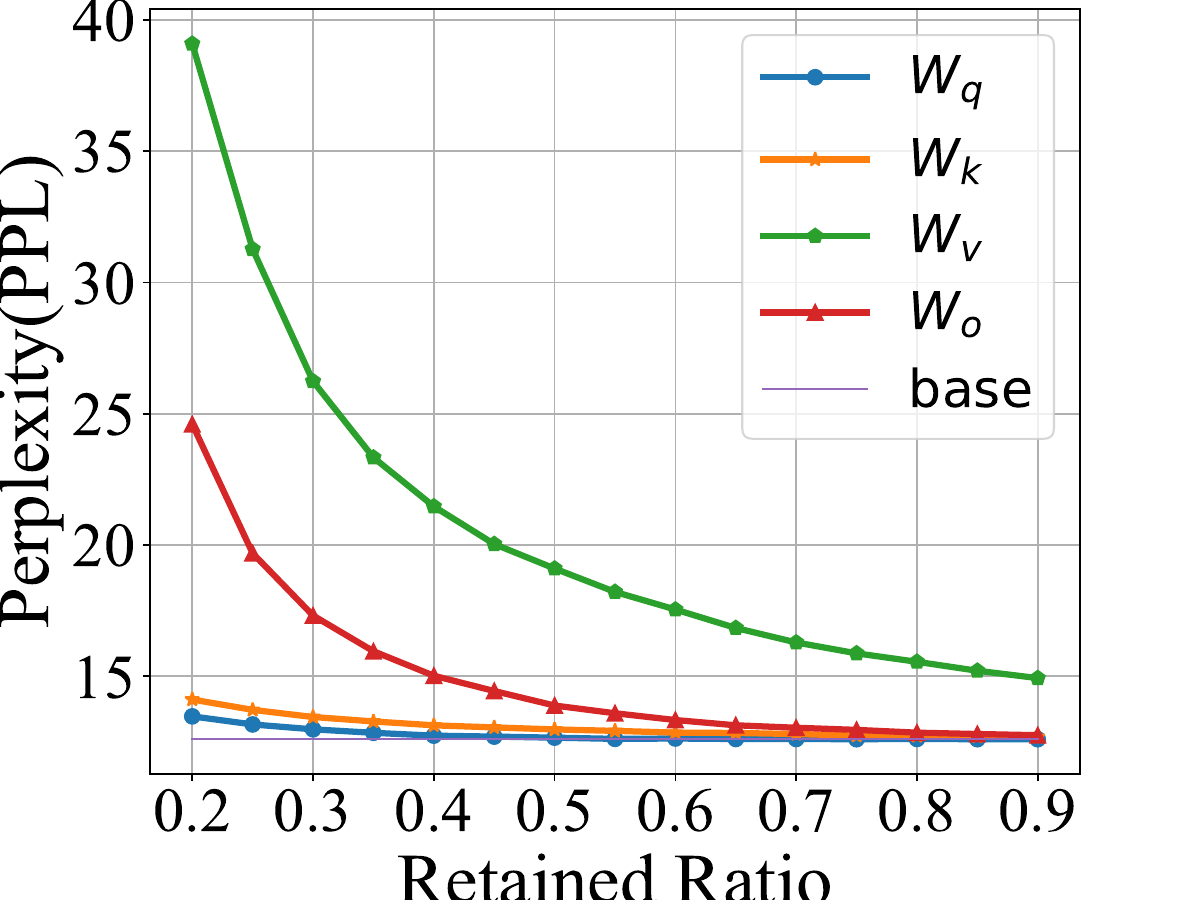}}
  \subfigure[PPL on PTB]{\includegraphics[width=0.23\textwidth]{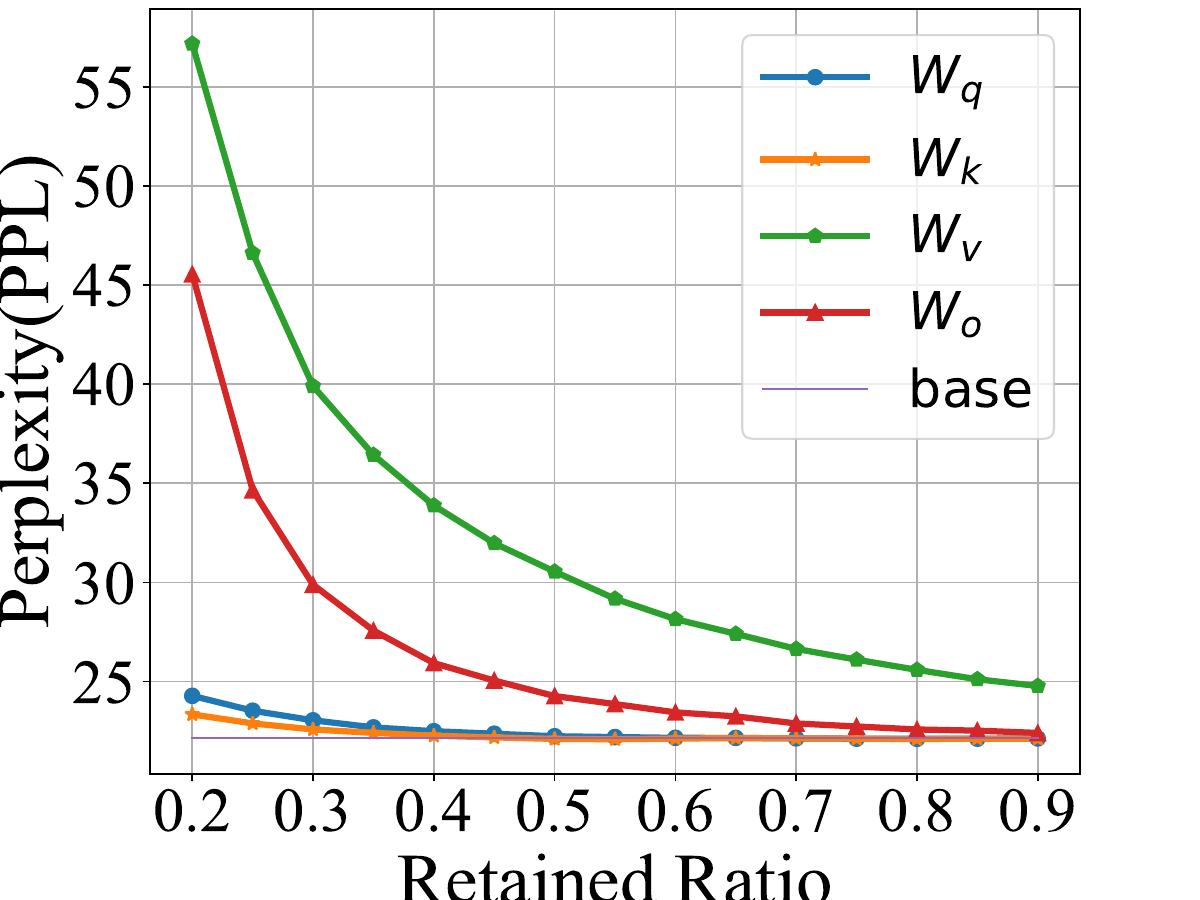}}
  \vskip -0.1in
  \caption{Different proportions are reserved for different weight matrices in MHA sub-layer. The parameters are increased by 0.5 each time from left to right. The perplexity (PPL) of the obtained model on Wikitext2 (left) and PTB (right) is present.} 
  \label{knowledge distribution}
\end{figure}

\textbf{Parameter Allocation.} In Table \ref{main struct}, \emph{we discover that the low-rank characteristic varies across different weight matrices in MHA sub-layer}. We further explore the impact of compressing these matrices on the final model performance. We separately compress the weight matrices in MHA sub-layer at different ratios and evaluate the  perplexity on  Wikitext2 and PTB datasets. The results are illustrated in  Fig. \ref{knowledge distribution}.
We observe that under the same compression ratio, the  model performance is better when compressing $\mathbf{W}_{q}$ and $\mathbf{W}_{k}$ matrices. This suggests that the distribution of knowledge in  $\mathbf{W}_{q}$ and $\mathbf{W}_{k}$ is more concentrated, thus requiring fewer parameters for preservation. 
This is consistent with the stronger low-rank property of $\mathbf{W}_{q}$ and $\mathbf{W}_{k}$ presented in Table \ref{low-rank-explore}.
In contrast, the knowledge distribution in  $\mathbf{W}_{v}$ and $\mathbf{W}_{o}$ matrices is more uniform, necessitating more parameters for storage.
Therefore, to achieve better performance within a limited parameter budget, more parameters should be allocated to  $\mathbf{W}_{v}$ and $\mathbf{W}_{o}$. In the experiments, we chose to allocate 75\% of the parameters to  $\mathbf{W}_{v}$  and $\mathbf{W}_{o}$ matrices, while allocating the remaining 25\% to  $\mathbf{W}_{q}$  and $\mathbf{W}_{k}$ matrices.

\begin{algorithm}[tb]
   \caption{LoRAP}
   \label{algo:AWSVD}
\begin{algorithmic}
\STATE {\bfseries Input:} The $i$-th layer $\mathcal{M}_{i}$ of model; input activation $\mathbf{X}_{in}^{i} \in \mathbb{R}^{N \times L \times d}$; retained ratio $p_{r}$ 

\FORALL{$\mathbf{W} \in \mathcal{M}_{i}$}
 \STATE Compute the $\mathbf{x}_{d_{in}}$ with input activation by Eq. (\ref{x_din});
\ENDFOR 
\FORALL{$\mathbf{W} \in  \mathcal{M}_{i}^{MHA}$}
    \STATE Compute the retained rank $r$ of $\mathbf{W}$;
    \STATE Compute the $\mathbf{L}$ and $\mathbf{R}$ by Eq. (\ref{FWSVD});
    \STATE Replace $\mathbf{W}$ by $\mathbf{L}_{r}\mathbf{R}_{r}$;
\ENDFOR \\
\FORALL{$\mathbf{W} \in  \mathcal{M}_{i}^{FFN}$}
     \STATE Compute the importance score $I(W_{ij})$ by Eq. (\ref{weight-importance});
     \STATE Compute the importance score of channel by Eq. (\ref{channel-importance});

\ENDFOR \\
\STATE Compute the importance score of group by Eq. (\ref{group-importance});
\STATE Prune $(1-p_{r})*100\%$ weights by Eq. (\ref{FFN MASK});
\STATE Compute input activation of next layer $\mathbf{X}^{i+1}_{in}=\mathbf{X}^{i}_{out}$;

\STATE {\bfseries Output:} The compressed layer $\mathcal{M}_{i}^{'}$; input activation of the $(i+1)$-th layer $\mathbf{X}^{i+1}_{in}$.
\end{algorithmic}
\end{algorithm}

\subsection{Gradient-Free Channel Pruning for FFN}
Both Table \ref{main struct} and  Fig. \ref{FFN visualition} indicate that the weights in  FFN are not suitable for low-rank approximation. Therefore, we opt to compress the FFN sub-layer with channel pruning. 
We estimate the  importance score $I(W_{ij})$ of weight $W_{ij}$  by Eq.(\ref{weight-importance}). Then, inspired by LoSparse \cite{Losparse}, we use the $\ell_2$ norm of the importance score of the weights $\mathbf{W}_{i,:}$ in the channel as the importance score of the $i$-th channel. The formulation is as follows:
\begin{align}\label{channel-importance}
    \Phi(\mathbf{W}_{i,:})=\|I(W_{i,1}), I(W_{i,2}), \cdots, I(W_{i,d_{in}})\|_{2}. 
\end{align}
Following  \cite{LLM-pruner}, we consider the dependencies between neurons during the pruning process. For example, as shown in Fig. \ref{main struct}, when pruning the $i$-th input channel of the down matrix $\mathbf{W}_{down}$, the corresponding output channels in the gate matrix $\mathbf{W}_{gate}$ and up matrix $\mathbf{W}_{up}$ should be pruned accordingly. 
Therefore, the interconnected channels are regarded as a group $\mathbf{W}^{group}_{i}$, i.e., 
\begin{equation}\label{group-def}
  \mathbf{W}^{group}_{i} = \{\mathbf{W}^{up}_{i,:}, \mathbf{W}^{gate}_{i,:}, \mathbf{W}^{down}_{:,i}\}.
\end{equation}
And the pruning is performed at the group level. We accumulate the channel importance to estimate the group importance as follows:
\begin{equation}\label{group-importance}
  C^{group}_{i}=\Phi(\mathbf{W}^{up}_{i,:}) +\Phi(\mathbf{W}^{gate}_{i,:})+\Phi(\mathbf{W}^{down}_{:,i}).
\end{equation}
\textbf{Retaining Least Important Weights.}
 Previous studies commonly prune the least important parts. However, \emph{we observe that the least important 1\% of parameters play a vital role in model performance.} 
 This phenomenon could be explained by Junk DNA Hypothesis \cite{JUNKDNA}, that is, certain less important weights actually encode crucial knowledge necessary for more difficult downstream tasks, and pruning these weights might severely destroy the model's performance. Therefore, while ensuring the pruning ratio remains unchanged, we retained the least important 1\% portion of weights.
The method for pruning is as follows:
\begin{equation}\label{FFN MASK}
\mathbf{W}^{group}_{i}=
\begin{cases}
  \mathbf{W}^{group}_{i}, & \text{if $C^{group}_{i}$ in top ($p_r*100$-1)\%, }\\
  \mathbf{W}^{group}_{i}, & \text{if $C^{group}_{i}$ in min 1\% },\\
  0, & \text{otherwise}.
\end{cases}
\end{equation}
Where $p_r$ represents the retained ratio. Finally, we summarize our algorithm in Algorithm \ref{algo:AWSVD}.

\subsection{Knowledge Recovery by LoRA}
In order to recovery the performance of the compressed model with limited data and computation, we adopt LoRA to  fine-tune the compressed model. We denote the pruned weight matrix in FFN as $\mathbf{W}_{f} \in \mathbb{R}^{d_{out}^{'} \times d_{in}^{'}} \ $, and the weight matrix $\mathbf{W}_{m}$ in MHA is decomposed into $\mathbf{L}$ and $\mathbf{R}$.
The update of weight matrix is denoted as $\triangle \mathbf{W}=\mathbf{AB}$. 
The forward computation can now be expressed as:
\begin{align}
     f_{FFN}(\mathbf{x})&=(\mathbf{W}_{f}+\triangle \mathbf{W})\mathbf{x}=(\mathbf{W}_{f}+\mathbf{A}_{f}\mathbf{B}_{f})\mathbf{x}\\
     f_{MHA}(\mathbf{x})&=(\mathbf{W}_{m}+\triangle \mathbf{W})\mathbf{x}=(\mathbf{L}\mathbf{R}+\mathbf{A}_{m}\mathbf{B}_{m})\mathbf{x}
\end{align}
During the training process, only training $\mathbf{A}$ and $\mathbf{B}$ can greatly reduces the computational workload and data requirements. After training, $\mathbf{A}$ and $\mathbf{B}$ can be directly merged with the compressed weight.
\section{Experiments}
\subsection{Experimental Settings}
\textbf{Benchmark LLMs.} To validate the effectiveness and generalization of our approach, we conducted experiments on LLaMA-1 \cite{LLaMA} , LLaMA-2 \cite{LLAMA-2} and Vicuna \cite{Vicuna} models. 
We conducted experiments on models with size 7B and 13B. This allows us to evaluate the effectiveness of our method across different model scales. In the main paper, we present the experimental results of LLaMA-1. More experimental results and analysis about LLaMA-2 and Vicuna models are present in Appendix \ref{appendix_B}.

\textbf{Calibration Data and Evaluation.}
For a more intuitive comparison with previous methods, we use the same calibration dataset and evaluation dataset as LLM-pruner \cite{LLM-pruner}. The calibration data is sampled from the BookCorpus \cite{bookcorpus}. We perform the zero-shot perplexity (PPL) evaluation on the WikiText2 \cite{wikitext2} and PTB datasets \cite{PTB}, which can roughly reflect the language capabilities of the model. To assess the zero-shot performance of the model in the task-agnostic setting, we follow LLaMA to perform zero-shot task classification on seven common sense reasoning datasets, including BoolQ \cite{BoolQ}, PIQA \cite{PIQA}, HellaSwag \cite{HellaSwag}, WinoGrande \cite{WinoGrande}, ARC-easy\cite{ARC-easy}, ARC-challenge \cite{ARC-challenge}, and OpenbookQA 
 \cite{OPENQA}.

\textbf{Implementation Details.}
During the model compression process, we randomly extract 128 samples from Bookcorpus as calibration data, and each sample is consisted of 128 tokens. 
During the knowledge recovery phase, we employ LoRA fine-tuning for two epochs on the cleaned Alpaca dataset \cite{alpaca}, which comprises approximately 50k samples. These experiments were conducted rigorously on A40 GPU (48G). 
After compression, we tested the compressed model on data segments containing 128 tokens in Wikitext2 and PTB and performed zero-shot classification tasks on commonsense reasoning datasets with lm-evaluation-harness \cite{lm-evaluation-harness}.

\textbf{Baselines.} We compare our proposal with the following well-performing structured compression methods:
\begin{itemize}
  \item LLM-pruner \cite{LLM-pruner} is the first structured pruning method applied to LLMs. Weights are organized into groups based on the dependency structure and then a portion of the groups with the lowest importance is pruned in one-shot.
  \item LoRAPrune \cite{LORApruner} uses the weights of LoRA to estimate the gradients of the model weights, which reduces both memory requirements and gradient computation. During the pruning process, the model weights are alternately updated and pruned until the model is pruned to the specified size.
  \item LoRAShear \cite{LORAshear} discovers minimally removal structures based on the dependency graphs and analyzes the knowledge distribution in layers. The weights  are then progressively pruned based on LoRA Half-Space Projected Gradient (LHSPG).
\end{itemize}

\begin{table*}[tb]
\caption{Zero-shot performance on LLaMA-7B models. At the same compression ratio, \textbf{`bold'} represents the best performance.}
  \centering
   \setlength\tabcolsep{4.5pt}
  \scriptsize
  \vskip 0.1in
   \begin{threeparttable}
  \begin{tabular}{cc|cc|cccccccc}
    \toprule
    \toprule
    Compression Ratio & Method  & WikiText2$\downarrow$ & PTB$\downarrow$ & BoolQ$\uparrow$ & PIQA$\uparrow$ & HellaSwag$\uparrow$ & WinoGrande$\uparrow$ & ARC-e$\uparrow$ & ARC-c$\uparrow$ & OBQA$\uparrow$ & Average$\uparrow$ \\
\midrule

   \multirow{2}{*}{Ratio=0\%} & LLaMA-7B & -     & -     & 76.5 & 79.8 & 76.1 & 70.1  & 72.8 & 47.6  & 57.2 & 68.59 \\
                              & LLaMA-7B* & 12.62 & 22.14 & 73.18& 78.35& 72.99& 67.01 & 67.45& 41.38 & 42.40 & 63.25 \\
    \midrule
    \multirow{4}{*}{\shortstack{Ratio=20\% \\ w/o tune}}
                                & LLM-pruner &19.09 & 34.21 & 57.06 & 75.68 & 66.80 & 59.83 & 60.94 & 36.52 & 40.00 & 56.69 \\
                                & LoRAPrune  &20.67 & 34.12 & 57.98 & 75.11 & 65.81 & 59.90 & \textbf{62.14} & 34.59 & 39.98 & 56.50 \\
                                & LoRAShear  &-     &  -    &   -   &    -  &   -   &   -   &  -    & -     &  -    &   -   \\
                                & LoRAP       &\textbf{15.69} & \textbf{25.86} & \textbf{71.93} & \textbf{76.44} & \textbf{69.98} & \textbf{65.90} & 60.56 & \textbf{38.48} & \textbf{40.40} & \textbf{60.53} \\
    \midrule
    \multirow{4}{*}{\shortstack{Ratio=20\% \\ w/ tune}}
                            & LLM-pruner &17.58 & 30.11 & 64.62 & 77.20 & 68.80 & 63.14 & 64.31 & 36.77 & 39.80 & 59.23 \\
                            & LoRAPrune  &16.80 & 28.75 & 65.62 & \textbf{79.31} & 70.00 & 62.76 & \textbf{65.87} & 37.69 & 39.14 & 60.05 \\
                            & LoRAShear  &-     &  -    & 70.17 & 76.89 & 68.69 & \textbf{65.83} & 64.11 & 38.77 & 39.97 & 60.63 \\
                            & LoRAP       &\textbf{16.35} & \textbf{27.06} & \textbf{72.94} & 76.93 & \textbf{70.90} & 65.75 & 64.31 &\textbf{ 39.93} & \textbf{41.20} & \textbf{61.71}  \\
    \midrule
    \midrule
    \multirow{4}{*}{\shortstack{Ratio=50\% \\ w/o tune}}
                                & LLM-pruner &112.44 & 255.38 & 52.32 & 59.63 & 35.64 & 53.20 & 33.50 & 27.22 & 33.40 & 42.13\\
                                & LoRAPrune  &121.96 & 260.14 & 51.78 & 56.90 & 36.76 & 53.80 & 33.82 & 26.93 & 33.10 & 41.87\\
                                & LoRAShear  &-      &  -     &   -   &    -  &   -   &   -   &  -    & -     &  -    &   -   \\
                                & LoRAP       &\textbf{56.96} & \textbf{87.71} & \textbf{57.80} & \textbf{63.82} & \textbf{46.96} & \textbf{57.30} & \textbf{40.36} & \textbf{27.73} & \textbf{36.80} & \textbf{47.25} \\

    \midrule
    \multirow{4}{*}{\shortstack{Ratio=50\% \\ w/ tune}}
                            & LLM-pruner &38.12 & 66.35 & 60.28 & 69.31 & 47.06 & 53.43 & 45.96 & 29.18 & 35.60 & 48.69\\
                            & LoRAPrune  &\textbf{30.12} & 50.30 & 61.88 & 71.53 & 47.86 & 55.01 & 45.13 & 31.62 & 34.98 & 49.71\\
                            & LoRAShear  &-     &  -    & 62.12 & \textbf{71.80} & 48.01 & 56.29 & 47.68 & \textbf{32.26} & 34.61 & 50.39 \\
                            & LoRAP       &30.90 & \textbf{48.84} & \textbf{63.00} & 69.64 & \textbf{54.42} & \textbf{58.41} & \textbf{51.94} & 32.00 & \textbf{35.80} & \textbf{52.17}   \\
    \bottomrule
    \bottomrule
  \end{tabular}
  \begin{tablenotes}[para,flushleft]
      \ * represents the evaluation version in LLM-Pruner \cite{LLM-pruner}. The average is calculated across seven common-sense reasoning datasets. 
  \end{tablenotes}
\vskip -0.1in
  \label{main-result}
    \end{threeparttable}
\end{table*}

\begin{table*}[tb]
\caption{Zero-shot performance on LLaMA-13B models. At the same compression ratio, \textbf{`bold'} represents the best performance. }
  \centering
  \setlength\tabcolsep{4.5pt}
  \scriptsize
  \vskip 0.1in
  \begin{threeparttable}
  \begin{tabular}{cc|cc|cccccccc}
    \toprule
    \toprule
    Compression Ratio & Method  & WikiText2$\downarrow$ & PTB$\downarrow$ & BoolQ$\uparrow$ & PIQA$\uparrow$ & HellaSwag$\uparrow$ & WinoGrande$\uparrow$ & ARC-e$\uparrow$ & ARC-c$\uparrow$ & OBQA$\uparrow$ & Average$\uparrow$ \\
    \midrule

     Ratio=0\% & LLaMA-13B* &11.58 & 20.24 & 68.47 & 78.89 & 76.24 & 70.09 & 74.58 & 44.54 & 42.00 & 64.97\\
    \midrule
    \multirow{2}{*}{\shortstack{Ratio=20\% \\ w/o tune}}
                                & LLM-pruner &16.01 & 29.28 & 67.68 & 77.15 & 73.41 & 65.11 & 68.35 & 38.40 & \textbf{42.40} & 61.79 \\
                                & LoRAP       &\textbf{13.48} & \textbf{23.57} & \textbf{73.94} & \textbf{77.31} & \textbf{74.93} & \textbf{69.69} & \textbf{70.79} & \textbf{40.44} & 41.40 & \textbf{64.07} \\

    \midrule
    \multirow{2}{*}{\shortstack{Ratio=20\% \\ w/ tune}}
                            & LLM-pruner &15.18 & 28.08 & 70.31 & 77.91 & 75.16 & 67.88 & \textbf{71.09} & 42.41 & \textbf{43.40} & 64.02 \\
                            & LoRAP       &\textbf{13.58} & \textbf{24.07} & \textbf{73.39} & \textbf{78.73} & \textbf{75.54} & \textbf{69.30} & 70.62 & \textbf{43.00} & 42.40 & \textbf{64.71}   \\
    \midrule
    \midrule
    \multirow{2}{*}{\shortstack{Ratio=50\% \\ w/o tune}}
                                & LLM-pruner &70.38 & 179.72 & 61.83 & 67.08 & 45.49 & 52.09 & 38.93 & 30.03 & 33.00 & 46.92  \\
                                & LoRAP       &\textbf{34.11} & \textbf{67.38}  & \textbf{65.81} & \textbf{70.46} & \textbf{57.80} & \textbf{59.04} & \textbf{51.22} & \textbf{30.88} & \textbf{34.40} & \textbf{52.80} \\

    \midrule
    \multirow{2}{*}{\shortstack{Ratio=50\% \\ w/ tune}}
                                & LLM-pruner &29.51 & 54.49 & 62.17 & 72.85 & 57.30 & 56.99 & 56.86 & 32.00 & 38.40 & 53.80  \\
                                & LoRAP       &\textbf{22.66} & \textbf{38.89} & \textbf{72.29} & \textbf{74.10} & \textbf{63.29} & \textbf{62.83} & \textbf{60.82} & \textbf{35.24} & \textbf{38.80} & \textbf{58.20}  \\
    \bottomrule
    \bottomrule
  \end{tabular}
   \begin{tablenotes}[para,flushleft]
      \ * represents the evaluation version in LLM-Pruner \cite{LLM-pruner}. The average is calculated across seven common-sense reasoning datasets. 
  \end{tablenotes}
  \label{13B-result}
  \vskip -0.2in
  \end{threeparttable}
\end{table*}


\subsection{Main Results}
\textbf{Zero-Shot Performance.} We compare our method with the baseline methods at different compression ratios. The experimental results are shown in Table \ref{main-result}. At different compression ratios, the perplexity of the compressed model on both WikiText2 and PTB datasets, is improved considerably, especially in settings that are not fine-tuned. At 50\% compression ratio without fine-tuning, we achieve perplexity of 56.96 and 87.71, respectively.
At a compression ratio of 20\% , our method achieves an average accuracy of 60.53\% on common sense reasoning datasets without fine-tuning, outperforming all previous structured pruning methods, and even comparable to the fine-tuned results of these methods. 
After fine-tuning, the performance was further improved, reaching the accuracy of 61.71\%, which is more than 1.0\% higher than baseline methods. At the 50\% compression ratio, our method outperformed baselines by 5\% in accuracy without fine-tuning, by approximately 2\% after fine-tuning. This demonstrates that across various compression ratios, our method consistently exhibits excellent performance.
Table \ref{13B-result} presents the performance of the compressed 13B model. We observe that our method outperforms the baseline methods significantly. Furthermore, the higher the compression ratio, the more evident the advantages of our method.  

\textbf{Statistics of the Compressed Model.}
Both structured pruning and matrix factorization can reduce computational complexity and model parameter count, directly. We use MACs to measure the computational complexity of the compressed model. Inference latency was tested in inference mode on the WikiText2 dataset with sentences composed of 64 tokens. To eliminate the influence of hardware, we compared our LoRAP with the LLM-Pruner \cite{LLM-pruner} on the same GPU device A40. The results are presented in the table \ref{test compressd model}. 
At the compression ratio of 50\%, the two methods yield compressed models with similar parameters amount.
LLM-Pruner shows notable reduction in computational complexity and 33.8\% inference acceleration. LORAP achieve similarly reduction in computational complexity and 31.25\% inference acceleration. 
 \begin{table}[tb]
\caption{The parameters, MACs, and inference latency of the baseline  and the compressed models.}
  \centering
  \setlength\tabcolsep{5pt}
  \scriptsize
   \vskip 0.1in
  \begin{tabular}{ccccc}
    \toprule
    Compression Ratio & Method  & Params & MACs & Latency \\
    \midrule
    Ratio=0\%  & Baseline & 6.74B	& 423.98 G &	65.79s \\
    \midrule
    \multirow{2}{*}{\shortstack{Ratio=50\%}} 
                                & LLM-Pruner &3.37B  & 206.59 G  & 43.55s (-33.80\%) \\
                                & LoRAP       & 3.37B & 208.40 G  & 45.23s (-31.25\%)  \\
    \bottomrule
  \end{tabular}
    \vskip -0.2in
  \label{test compressd model}
\end{table}

\subsection{Ablation Study}
\textbf{Retention of the Least Important Weights.} In order to investigate the impact of retaining the minimum importance weights in the pruning of the FFN sub-layer, we conducted ablation experiments. We regard the non-retention compression as the baseline and retain different proportions of the minimum importance weights at two compression ratios. The experimental results are presented in Table \ref{7B-least-saved}. 
The result shows that the adoption of retention will bring noticeable performance improvement, with a more pronounced enhancement at higher compression ratios. At the compression ratio of 50\%, the model with  retention achieve a reduction of approximately 15\% in perplexity on Wikitext2 and PTB, along with a 1\% increase in average accuracy on the seven reasoning datasets.  Furthermore, it is worth noting that only the lowest important 1\% of the parameters effectively improve the model performance, and retaining a higher percentage of parameters does not result in further significant improvement. 

\begin{table}[tb]
\caption{The results of retaining different proportions of the least importance weights in the FFN sub-layer.} 
  \centering
  \setlength\tabcolsep{5.5pt}
  \scriptsize
  \vskip 0.1in
  \begin{tabular}{cc|ccc}
    \toprule
    Compression Ratio & Retained Ratio & WikiText2$\downarrow$ & PTB$\downarrow$ & Average$\uparrow$ \\
    \midrule
    \multirow{3}{*}{\shortstack{Ratio=20\%}} 
                                & 0\% &16.91 & 27.96  & 60.02 \\
                                & 1\% &15.69 & 25.86  & \textbf{60.53} \\
                                & 2\% &\textbf{15.61} &	\textbf{25.80} & 60.23  \\
    \midrule
    \multirow{3}{*}{\shortstack{Ratio=50\%}} 
                                & 0\% &67.53 & 102.93  & 46.19 \\
                                & 1\% &\textbf{56.96} &	87.71 & \textbf{47.25}    \\
                                & 2\% &57.88 & \textbf{86.44}    & 46.78 \\
    \bottomrule
  \end{tabular}
 \vskip -0.3in
  \label{7B-least-saved}
\end{table}

\textbf{Parameter Allocation in MHA Sub-Layer.}
Table \ref{low-rank-explore} and Fig. \ref{knowledge distribution} demonstrate that different modules in the MHA sub-layer exhibit  varying degrees of low rank properties.
Given the parameter budget constraint, it is  reasonable to allocate more parameters to  $\mathbf{W}_{v}$  and $\mathbf{W}_{o}$ matrices since both of them possess poorer low-rank characteristic.  The $\mathbf{W}_{q}$  and $\mathbf{W}_{k}$ matrices are treated as a group, while the $\mathbf{W}_{v}$ and $\mathbf{W}_{o}$ matrices are regarded as another group.
The number of parameters differs between the two groups.
To investigate the impact of parameter allocation, we adjusted the parameter ratio between two groups, without altering the pruning of the FFN sub-layer. The results of the different parameter allocations are presented in  Table \ref{allocate ablate}, where ratio denotes the parameter ratio of ($\mathbf{W}_{q}$  + $\mathbf{W}_{k}$) $:$ ($\mathbf{W}_{v}$  + $\mathbf{W}_{o}$). It can be seen that  with the same number of parameters, the performance of the compression model is remarkably improved after adopting the parameter allocation in MHA sub-layer. Moreover, the parameter ratio of 1:3 achieves superior  overall performance. Therefore, we choose the parameter ratio of 1:3 for regular experiments.
\begin{table}[tb]
\caption{Different parameter allocation in the MHA sub-layer. }
  \centering
  \setlength\tabcolsep{8.3pt}
  \scriptsize
  \vskip 0.1in
  \begin{tabular}{cc|ccc}
   \toprule
    Compression Ratio & Ratio  & WikiText2$\downarrow$ & PTB$\downarrow$ & Average$\uparrow$ \\
    \midrule
    \multirow{5}{*}{\shortstack{Ratio=50\%}} 
                                & 1:1      & 88.81 & 127.66 & 40.70   \\
                                & 1:2      & 60.39 & 92.17 & 45.32   \\
                                & 1:2.5    & 56.69 & 87.08 & 45.30   \\
                                & 1:3      & 56.38 & \textbf{87.37} & \textbf{47.25 }  \\
                                & 1:3.5    & \textbf{55.43} & 87.55 & 46.84   \\
    \bottomrule
  \end{tabular}
  \vskip -0.1in
  \label{allocate ablate}
\end{table}

\begin{table}[tb]
\caption{The  results of three different aggregation strategies for channel importance score. }
  \centering
  \setlength\tabcolsep{8pt}
  \scriptsize
    \vskip 0.1in
  \begin{tabular}{cl|ccc}
    \toprule
    Compression Ratio & Norm  & WikiText2$\downarrow$ & PTB$\downarrow$ & Average$\uparrow$ \\
    \midrule
    \multirow{3}{*}{\shortstack{Ratio=20\%}} 
                                & $\ell_1$      & 15.80 & 26.04 & \textbf{60.94}  \\
                                & $\ell_2$      & 15.69 & \textbf{25.86} & 60.53  \\
                                & $\ell_{\infty}$ & \textbf{15.68} & 26.08 & 59.89  \\
    \midrule
    \multirow{3}{*}{\shortstack{Ratio=50\%}} 
                                & $\ell_1$      & 58.58 & 92.29 & 45.42  \\
                                & $\ell_2$       & \textbf{56.96} & \textbf{87.71} & \textbf{47.25}  \\
                                & $\ell_{\infty}$ & 71.91 & 103.06 & 44.09  \\
    \bottomrule
  \end{tabular}
  \label{abalate norm}
  \vskip -0.1in
\end{table}

\begin{table}[!tb]
\caption{Comparison of different structured compression methods in FFN  and MHA sub-layers.  * denotes adopting the same parameter allocation scheme as our AWSVD.}
  \centering
  \setlength\tabcolsep{3.8pt}
  \scriptsize
  \vskip 0.1in
  \begin{tabular}{ccc|cc}
    \toprule
     Sub-layer & Compression Ratio & Method  & WikiText2$\downarrow$ & PTB$\downarrow$ \\
    \midrule
    \multirow{4}{*}{\shortstack{FFN}} & \multirow{4}{*}{\shortstack{Ratio=50\%}}
                                                & SVD    &14554.00 & 19269.28   \\
                                           &  & AFM    &126.13 & 274.53   \\
                                           &  & Our AWSVD  &225.08 & 319.40   \\
                                           &  & Our Channel Prune  &\textbf{32.64}  &\textbf{ 61.51}    \\
    \midrule
    \multirow{6}{*}{\shortstack{MHA}}  & \multirow{6}{*}{\shortstack{Ratio=50\%}}
                                                & AFM    &33.90 & 83.84     \\
                                            &  & AFM*   &25.09 & 58.80    \\
                                            &  & SVD    &345.16 & 778.82     \\
                                            &  & SVD*   &76.54 & 196.32     \\
                                            &  & Head Prune  &458.93 & 414.61     \\
                                            &  &Our AWSVD  &\textbf{19.49} & \textbf{29.66 }   \\
    \bottomrule
  \end{tabular}
  \label{ablate AFM}
  \vskip -0.1in
\end{table}

\textbf{Aggregation Strategies for Channel Importance Score.} 
Through the input activations, we estimate the importance score of weight $ I(W_{ij})$ by Eq. (\ref{weight-importance}). However, during the pruning process, we rely on channel importance $\Phi(\mathbf{W}_{i,:})$. Therefore, we need to choose a suitable aggregation strategy based on the importance score of weights to estimate the channel importance. There are typically three aggregation strategies: (1) $\ell_1$ norm aggregation: $\Phi(\mathbf{W}_{i,:})=\sum_{j}^{d_{in}}I(W_{ij})$; (2) $\ell_2$ norm aggregation: $\Phi(\mathbf{W}_{i,:})=(\sum_{j}^{d_{in}}I(W_{ij})^2)^{1/2}$; (3) $\ell_{\infty}$ norm aggregation: $\Phi(\mathbf{W}_{i,:}) = \mathrm{Max}(I(W_{ij}))$. The results of  three aggregation strategies are shown in  Table \ref{abalate norm}. At the compression ratio of 20\%, the three methods obtain comparable performance. But at the compression ratio of 50\%,
$\ell_2$ norm exhibits clearly superior performance.
Therefore, we adopt the $\ell_2$ norm  for channel importance score in the experiments.

\textbf{Comparison of Structured Compression Methods.} 
In this part, we further compare our proposal with other low-rank approximation and pruning methods.
 Atomic Feature Mimicking (AFM) \cite{AFM} was successfully used in LORD \cite{LORD} to compress a 16B code model and SVD can also be used for large models without relying on data. Structured pruning in MHA is usually attention head pruning \cite{LLM-pruner,LORApruner}. We compare our method with them in two sub-layers respecively. In the MHA sub-layer, for a fair comparison, we used the same parameter allocation scheme as our AWSVD for AFM and SVD. The results are shown in Table \ref{ablate AFM}.
It can be observed that in the FFN sublayer, channel pruning surpasses various low-rank approximation methods by a large margin. However, in the MHA sub-layer, the attention head pruning is generally worse than low-rank approximation.  Besides, we discover that the proposed parameter allocation scheme not only works for our AWSVD but also  improves the performance of AFM and SVD, which confirms its excellent generalization. Lastly, our proposed AWSVD method is obviously superior to AFM and SVD,  validating the effectiveness of our input activation weighted mechanism.

\vspace{0.05cm}

\section{Conclusion}
Different from the existing works that compress the Transformer modules of  LLMs with the same way, this work proposes a mixed structured compression model named LoRAP, which employs low-rank approximation and structured pruning separately for different sub-layers of Transformer.  LoRAP draws inspiration from our observation that the MHA sub-layer presents noticeable low-rank pattern, while the FFN sub-layer does not. For the MHA sub-layer, a weighted low-rank approximation method is proposed, which adopts input activation as the weighted matrix and allocates the  parameters  according to the low-rank degrees. For the FFN sub-layer, a gradient-free structured pruning method is devised. The  results indicate that under multiple compression ratios, LoRAP is superior to previous structured compression methods with or without fine-tuning. 
\section{Impact Statements}
This paper presents work whose goal is to advance the field of Machine Learning. There are many potential societal consequences of our work, none which we feel must be specifically highlighted here.




\nocite{langley00}

\bibliography{example_paper}
\bibliographystyle{icml2024}

\newpage
\appendix
\onecolumn

\section{Ablation Studies}
\subsection{ Impact of Calibration Data}
\begin{figure}[h]
  \centering
\includegraphics[width=0.40\textwidth]{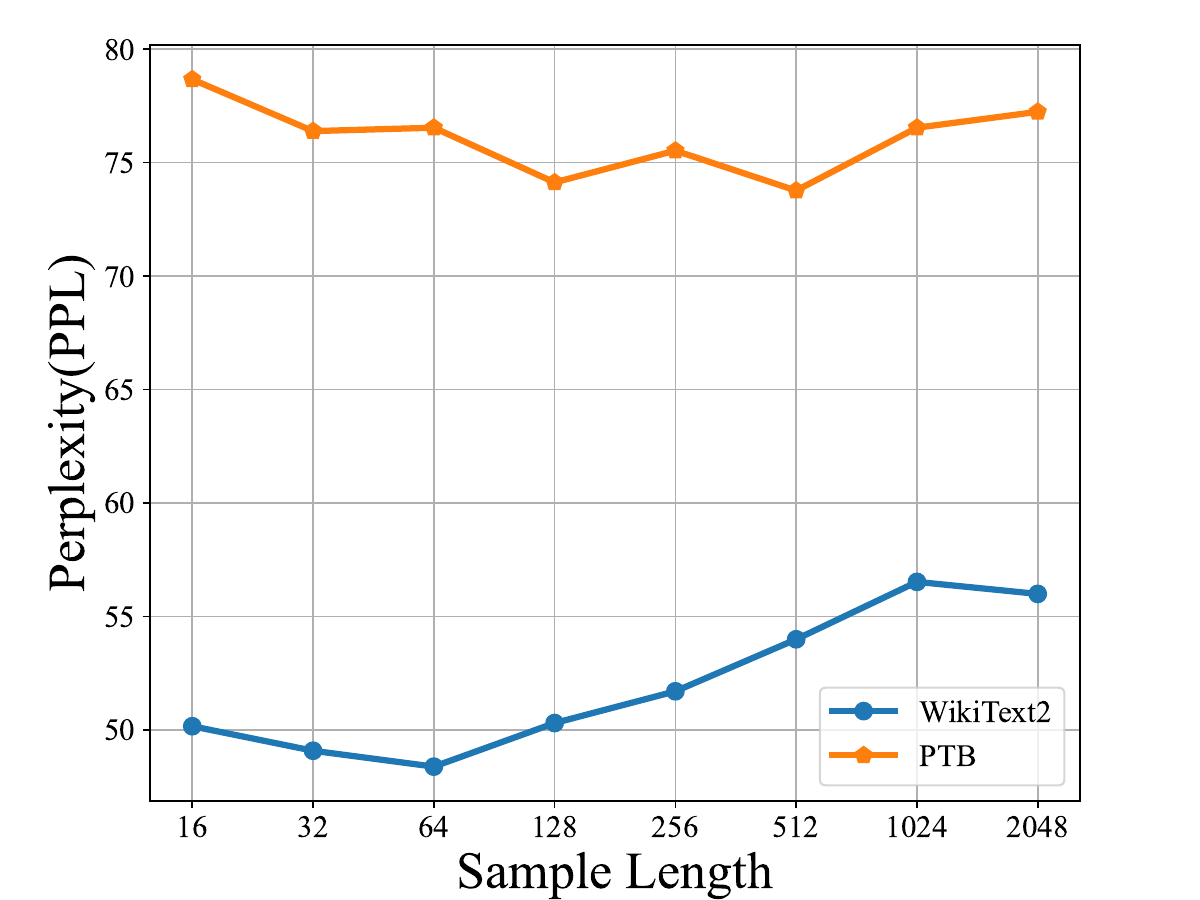}
  \hspace{0.1 in} 
 \includegraphics[width=0.40\textwidth]{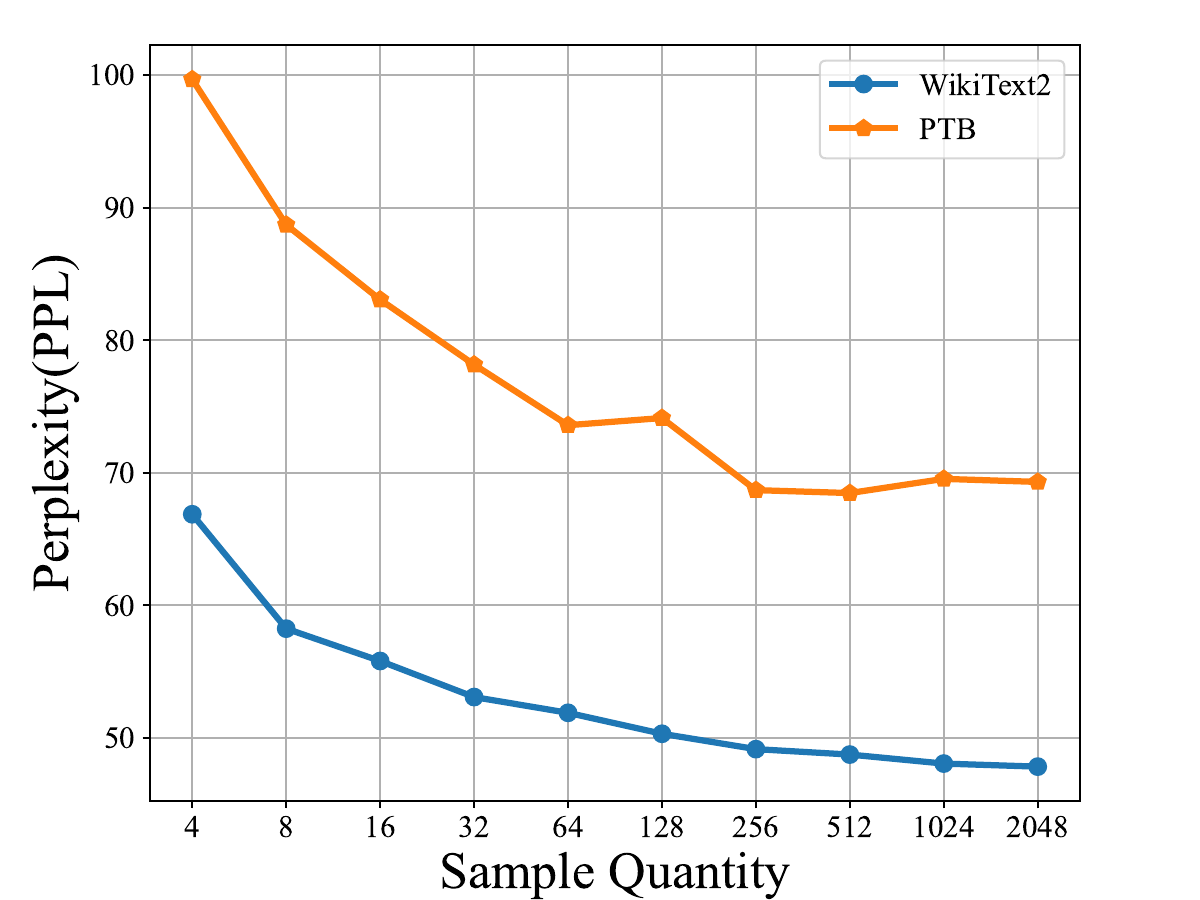}
   \vskip -0.1in
  \caption{As the quantity and length of calibration data increases, the evaluation results of the compressed model on WikiText2 and PTB.}
  \label{samples number}
\end{figure}

We explore the influence of both the length and the quantity of calibration data on model compression.  To meet the length requirements for sampling, we sampled the calibration data from the C4 dataset. Under the configuration where the sample length is fixed at 128 tokens, we gradually increase the sample quantity from 4 to 2048. Similarly, maintaining the sample quantity of 128, we progressively increase the sample length from 16 to 2048 tokens. The results are shown in the figure \ref{samples number}. 
The results indicate that with the increase in the quantity of calibration data, the perplexity of the model decreases on both datasets. This suggests that augmenting the quantity of calibration data can effectively enhance the performance of the compressed model.
However, with the increase in sample length, the perplexity of the model decreases initially and then increases on both datasets. The above analysis indicate that increasing the quantity of calibration data and selecting an appropriate length can effectively improve the performance of the compressed model.
\vskip -0.1 in
\vskip -0.1 in
\subsection{Sensitivity to Random Seeds}
In this part, we investigate the sensitivity of our algorithm to randomness. We conducted 10 runs with different random seeds at compression ratios of 20\% and 50\%, obtaining the results on  WikiText2 of 15.734 ± 0.073 (mean/standard deviation) and 49.116 ± 0.947, respectively. These findings indicate a strong robustness of the proposed method to variations in random seeds.
\section{Additional Models} \label{appendix_B}
\begin{figure}[h]
  \centering
  \subfigure[LLaMA-7B]{\includegraphics[width=0.3\textwidth]{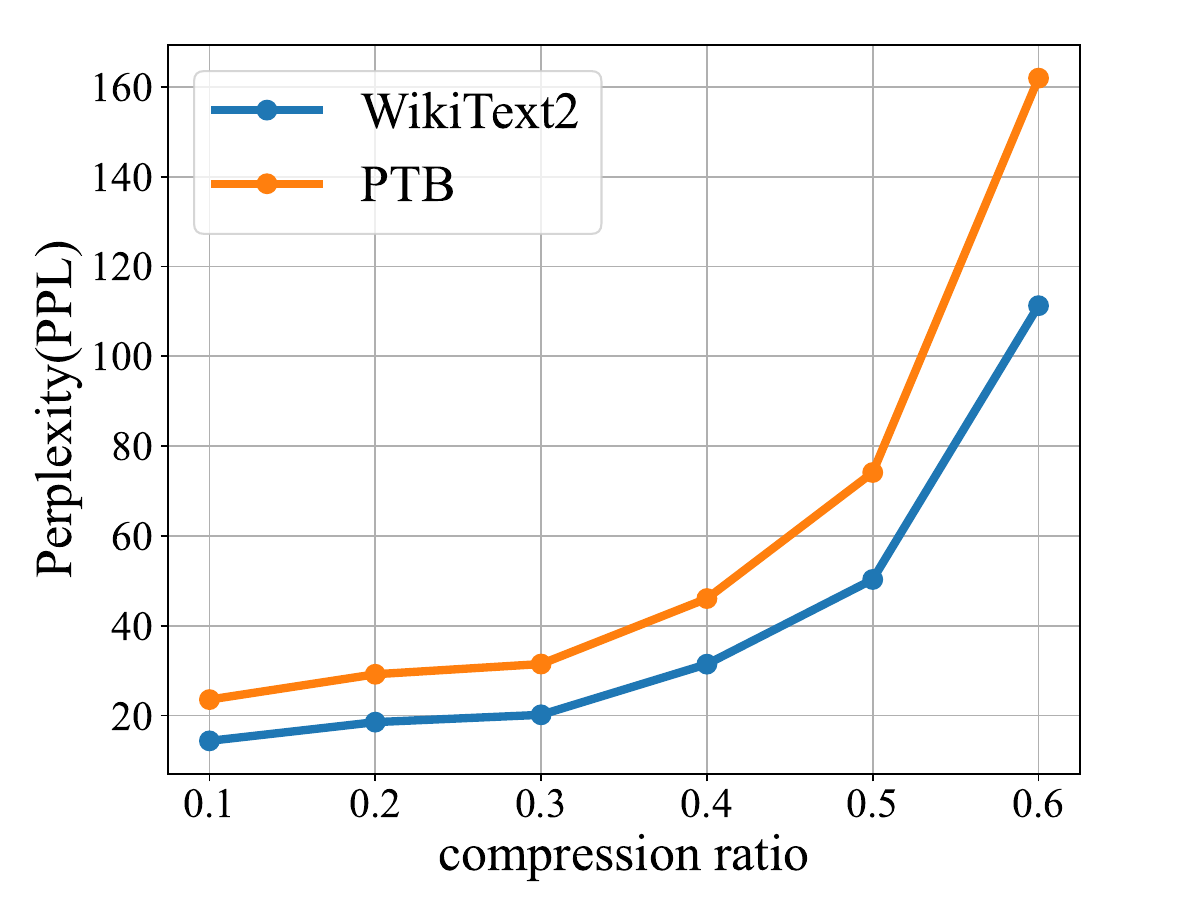}}
  \hspace{5pt} 
  \subfigure[LLaMA-13B]{\includegraphics[width=0.3\textwidth]{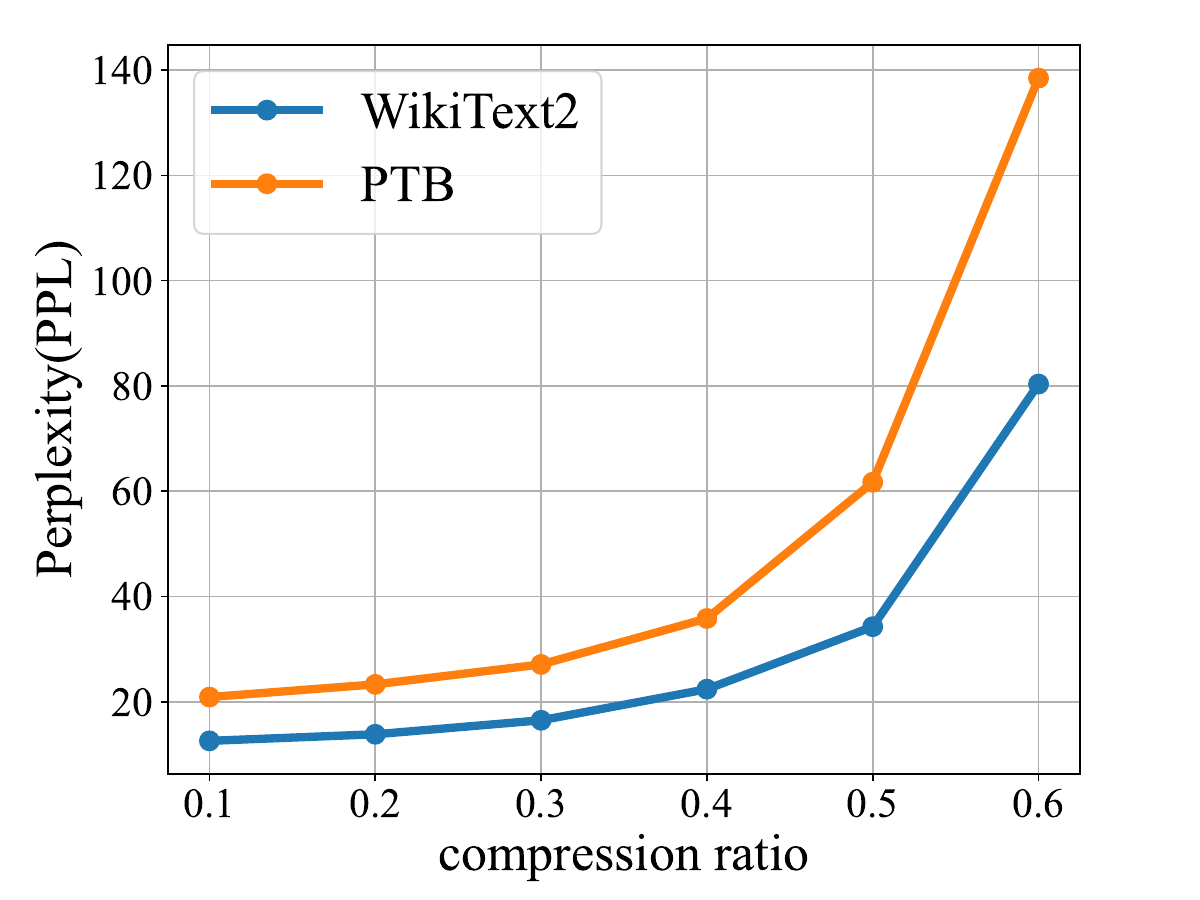}}
  \hspace{5pt} 
  \subfigure[LLaMA-30B]{\includegraphics[width=0.3\textwidth]{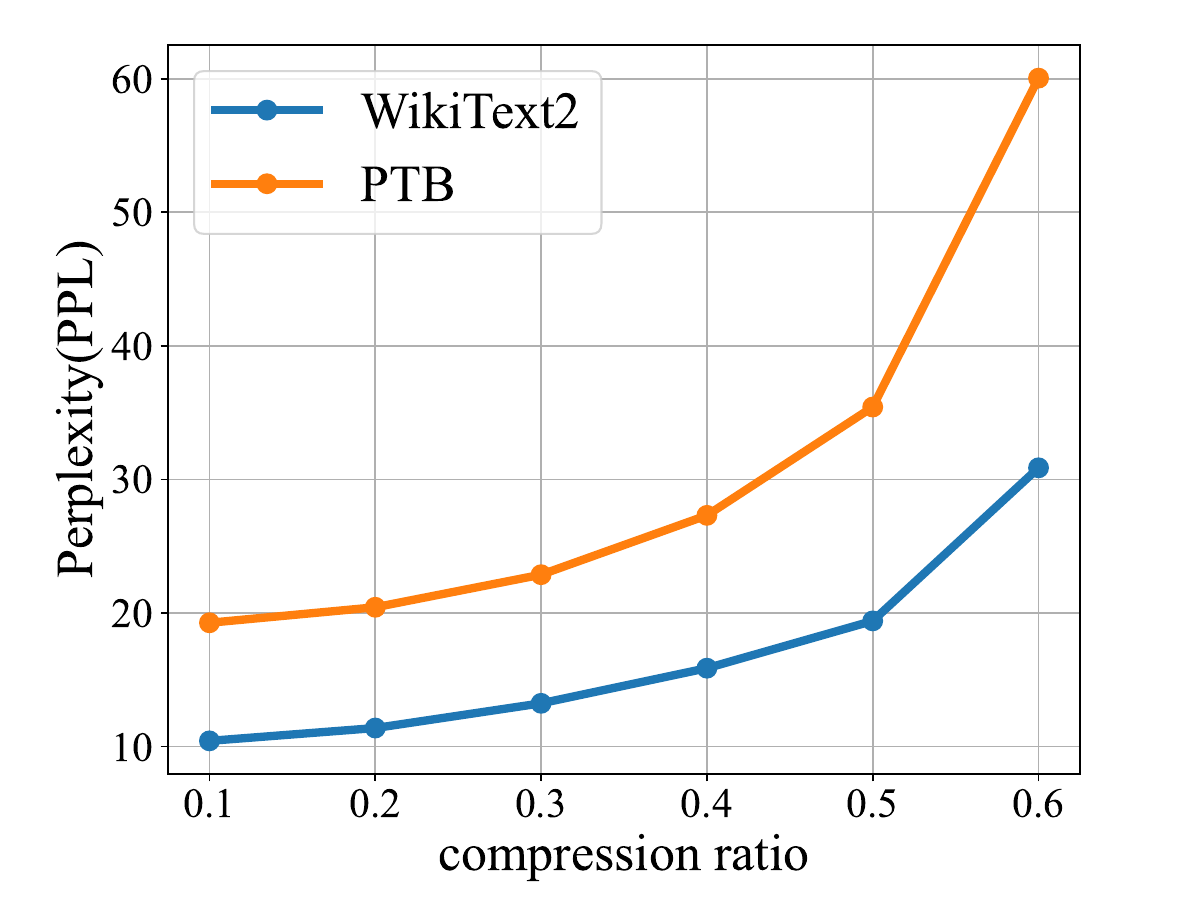}}
  \vskip -0.1in
  \caption{The performance of the compressed model with more compression ratios and model sizes}
  \label{LLaMA-1-all}
\end{figure}

We compress the 7B, 13B, and 30B LLaMA models under six different compression ratios, and the evaluation of the compressed model on WikiText2 and PTB datasets are illustrated in Fig. \ref{LLaMA-1-all}. It can be observed that, at the same compression ratio, larger models preserve performance more comprehensively. 
This indicates that larger models contain more redundant parameters and possess a larger compression space.
\vskip -0.1in

\newpage
\begin{table}[!htb]
\caption{The performance of the compressed Vicuna-7B models.}
  \centering
  \scriptsize
  \vskip 0.1in
  \begin{tabular}{cc|cc|cccccccc}
    \toprule
    \toprule
    Method & Pruning Ratio & WikiText2$\downarrow$ & PTB$\downarrow$ & BoolQ$\uparrow$ & PIQA$\uparrow$ & HellaSwag$\uparrow$ & WinoGrande$\uparrow$ & ARC-e$\uparrow$ & ARC-c$\uparrow$ & OBQA$\uparrow$ & Average$\uparrow$ \\
    \midrule
    Base* & Ratio=0\% & 16.24 & 60.78 & 75.69 & 77.09 & 71.04 & 67.88 & 68.98 & 39.93 & 42.40 & 63.29   \\
    \midrule
    \multirow{5}{*}{\shortstack{ w/o tune}}
                                & Ratio=10\%  &17.86 & 65.62 & 78.04 & 76.61 & 70.12 & 66.14 & 67.21 & 38.40 & 41.40 & 62.56   \\
                                & Ratio=20\%  &20.07 & 71.75 & 76.42 & 76.38 & 68.31 & 64.96 & 65.82 & 37.29 & 38.60 & 61.11   \\
                                & Ratio=30\%  &24.94 & 88.58 & 72.75 & 74.86 & 65.60 & 63.77 & 62.12 & 34.98 & 39.00 & 59.01    \\
                                & Ratio=40\%  &42.60 & 148.49 & 66.27 & 68.82 & 56.75 & 58.48 & 50.88 & 31.83 & 36.60 & 52.80    \\
                                & Ratio=50\%  &84.17 & 272.01 & 63.88 & 63.66 & 46.59 & 56.35 & 39.81 & 27.82 & 35.20 & 47.62    \\
                               
    \midrule
    \multirow{6}{*}{\shortstack{ w/tune}}
                                & Ratio=10\%  &15.87 & 58.04 & 77.65 & 77.53 & 70.46 & 67.25 & 71.38 & 40.02 & 41.40 & 63.67    \\
                                & Ratio=20\%  &17.33 & 62.18 & 75.81 & 76.77 & 68.39 & 65.04 & 70.08 & 39.33 & 39.20 & 62.09   \\
                                & Ratio=30\%  &19.63 & 70.70 & 73.91 & 75.84 & 65.45 & 63.77 & 66.96 & 35.84 & 39.00 & 60.11     \\
                                & Ratio=40\%  &24.53 & 86.97 & 70.95 & 73.01 & 60.03 & 62.19 & 59.01 & 33.96 & 38.20 & 56.76     \\
                                & Ratio=50\%  &30.68 & 104.49 & 68.65 & 69.86 & 54.63 & 56.84 & 54.59 & 31.14 & 38.40 & 53.44    \\
    \bottomrule
    \bottomrule
  \end{tabular}
\vskip -0.2in
  \label{vicuna-7B}
\end{table}

\begin{table}[!htb]
\caption{The performance of the compressed Vicuna-13B models.}
  \centering
  \scriptsize
  \vskip 0.1in
  \begin{tabular}{cc|cc|cccccccc}
    \toprule
    \toprule
    Method & Pruning Ratio & WikiText2$\downarrow$ & PTB$\downarrow$ & BoolQ$\uparrow$ & PIQA$\uparrow$ & HellaSwag$\uparrow$ & WinoGrande$\uparrow$ & ARC-e$\uparrow$ & ARC-c$\uparrow$ & OBQA$\uparrow$ & Average$\uparrow$ \\
    \midrule

    Base* & Ratio=0\% & 13.51 & 56.43 & 76.51 & 78.73 & 74.63 & 69.06 & 72.35&44.80 & 41.00 & 65.30     \\
    \midrule
    \multirow{5}{*}{\shortstack{ w/o tune}}
                                & Ratio=10\%  &15.14 & 60.41 & 77.28 & 78.19 & 73.99 & 68.59 & 71.63 & 43.17 & 41.40 & 64.89      \\
                                & Ratio=20\%  &17.08 & 66.52 & 80.58 & 77.91 & 73.80 & 68.59 & 71.84 & 42.92 & 40.60 & 65.18      \\
                                & Ratio=30\%  &20.46 & 78.23 & 79.79 & 77.09 & 72.27 & 66.46 & 69.74 & 42.75 & 40.00 & 64.01     \\
                                & Ratio=40\%  &26.92 & 108.53 & 75.23 & 72.96 & 66.54 & 61.56 & 63.89 & 39.16 & 39.60 & 59.85      \\
                                & Ratio=50\%  &42.50 & 182.75  & 71.90 & 69.91 & 57.92 & 61.64 & 51.77 & 32.85 & 37.60 & 54.80     \\    
    \midrule
    \multirow{6}{*}{\shortstack{ w/tune}}
                                & Ratio=10\%  &12.98 & 57.62 & 80.73 & 79.16 & 74.24 & 69.77 & 73.83 & 44.54 & 41.20 & 66.21    \\
                                & Ratio=20\%  &14.06 & 61.28 & 80.12 & 78.51 & 72.79 & 66.06 & 73.19 & 43.17 & 40.60 & 64.92    \\
                                & Ratio=30\%  &15.88 & 67.28 & 79.45 & 77.09 & 71.07 & 65.90 & 70.58 & 40.87 & 40.80 & 63.68    \\
                                & Ratio=40\%  &19.14 & 80.81 & 75.29 & 75.73 & 67.69 & 62.19 & 67.17 & 40.44 & 39.40 & 61.13      \\
                                & Ratio=50\%  &23.35 & 94.99 & 73.43 & 73.34 & 62.39 & 61.01 & 63.89 & 36.77 & 37.60 & 58.35     \\
    \bottomrule
    \bottomrule
  \end{tabular}
\vskip -0.2in
  \label{vicuna-13B}
\end{table}

\begin{table}[!htb]
\caption{The performance of the compressed LLaMA2-7B models.}
  \centering
  \scriptsize
  \vskip 0.1in
  \begin{tabular}{cc|cc|cccccccc}
    \toprule
    \toprule
   Method & Pruning Ratio & WikiText2$\downarrow$ & PTB$\downarrow$ & BoolQ$\uparrow$ & PIQA$\uparrow$ & HellaSwag$\uparrow$ & WinoGrande$\uparrow$ & ARC-e$\uparrow$ & ARC-c$\uparrow$ & OBQA$\uparrow$ & Average$\uparrow$ \\
    \midrule

    Base* & Ratio=0\% & 12.19 & 48.35 & 71.04 & 78.40 & 72.96 & 67.17 & 69.32 & 40.53 & 40.80 & 62.89  \\
    \midrule
    \multirow{5}{*}{\shortstack{ w/o tune}}
                                & Ratio=10\%  &13.23 & 51.23 & 72.20 & 77.86 & 71.67 & 66.54 & 66.17 & 39.59 & 39.80 & 61.98   \\
                                & Ratio=20\%  &15.02 & 58.44 & 69.24 & 76.39 & 69.15 & 65.11 & 61.99 & 35.58 & 38.60 & 59.44    \\
                                & Ratio=30\%  &18.58 & 73.08 & 65.93 & 74.70 & 64.76 & 62.90 & 54.00 & 32.76 & 35.80 & 55.84    \\
                                & Ratio=40\%  &30.94 & 133.39 & 62.11 & 67.52 & 55.98 & 58.72 & 47.35 & 30.97 & 36.00 & 51.24    \\
                                & Ratio=50\%  &60.89 & 282.22 & 61.86 & 62.23 & 43.98 & 55.41 & 38.51 & 27.65 & 33.00 & 46.09     \\
                               
    \midrule
    \multirow{5}{*}{\shortstack{ w/tune}}
                                & Ratio=10\%  &13.23 & 52.87 & 75.20 & 78.89 & 72.76 & 67.80 & 68.52 & 40.53 & 40.60 & 63.47    \\
                                & Ratio=20\%  &14.67 & 57.52 & 70.89 & 78.13 & 69.93 & 65.67 & 65.99 & 38.48 & 39.60 & 61.24  \\
                                & Ratio=30\%  &16.84 & 66.24 & 69.60 & 76.71 & 66.75 & 62.98 & 60.61 & 35.49 & 37.80 & 58.56   \\
                                & Ratio=40\%  &21.46 & 82.51 & 67.22 & 73.83 & 61.57 & 61.01 & 56.78 & 32.33 & 37.20 & 55.71   \\
                                & Ratio=50\%  &26.26 & 101.22& 63.27 & 70.78 & 55.14 & 57.85 & 52.15 & 30.97 & 36.00 & 52.31  \\
    \bottomrule
    \bottomrule
  \end{tabular}
\vskip -0.25in
\label{LLaMA2-7B}
\end{table}

\begin{table}[!htb]
\caption{The performance of the compressed LLaMA2-13B models.}
  \centering
  \scriptsize
  \vskip 0.1in
  \begin{tabular}{cc|cc|cccccccc}
    \toprule
    \toprule
   Method & Pruning Ratio & WikiText2$\downarrow$ & PTB$\downarrow$ & BoolQ$\uparrow$ & PIQA$\uparrow$ & HellaSwag$\uparrow$ & WinoGrande$\uparrow$ & ARC-e$\uparrow$ & ARC-c$\uparrow$ & OBQA$\uparrow$ & Average$\uparrow$ \\
    \midrule

    Base* & Ratio=0\% & 10.98 & 54.42 & 69.02 & 78.72 & 76.59 & 69.53 & 73.27 & 44.20 & 42.00 & 64.76  \\
    \midrule
    \multirow{5}{*}{\shortstack{ w/o tune}}
                                & Ratio=10\%  &12.13 & 57.02 & 73.49 & 79.05 & 75.94 & 69.22 & 73.23 & 43.60 & 42.40 & 65.28  \\
                                & Ratio=20\%  &13.33 & 62.64 & 74.62 & 78.78 & 74.72 & 67.56 & 71.55 & 42.15 & 42.20 & 64.51  \\
                                & Ratio=30\%  &16.11 & 77.39 & 74.46 & 77.04 & 72.09 & 65.27 & 67.30 & 39.76 & 40.80 & 62.39   \\
                                & Ratio=40\%  &21.19 &115.39 & 65.87 & 72.80 & 65.48 & 63.77 & 58.08 & 35.24 & 39.40 & 57.23   \\
                                & Ratio=50\%  &34.43 &204.51 & 66.57 & 68.61 & 54.62 & 59.83 & 47.85 & 29.95 & 35.20 & 51.80   \\
                               
    \midrule
    \multirow{6}{*}{\shortstack{ w/tune}}
                                & Ratio=10\%  &11.94 & 57.64 & 76.70 & 79.92 & 76.61 & 69.22 & 75.21 & 44.37 & 42.80 & 66.40   \\
                                & Ratio=20\%  &12.98 & 62.30 & 77.25 & 79.16 & 75.31 & 67.48 & 73.95 & 43.94 & 41.80 & 65.56  \\
                                & Ratio=30\%  &14.95 & 71.45 & 76.88 & 77.69 & 73.87 & 66.22 & 71.63 & 41.55 & 41.80 & 64.23    \\
                                & Ratio=40\%  &17.66 & 87.42 & 72.26 & 75.79 & 69.21 & 64.40 & 65.28 & 39.68 & 40.80 & 61.06    \\
                                & Ratio=50\%  &22.05 &104.23 & 69.88 & 74.16 & 63.68 & 61.09 & 61.11 & 36.60 & 39.00 & 57.93    \\
    \bottomrule
    \bottomrule
  \end{tabular}
\vskip -0in
  \label{LLaMA2-13B}
\end{table}

\newpage
Furthermore, we compress the Vicuna-7B, Vicuna-13B, LLaMA2-7B, LLaMA2-13B, models under five different compression ratios.
Moreover, the same knowledge recovery method as described in the main paper was employed to restore the performance of the compressed models. 
The evaluation results are presented in Table \ref{vicuna-7B}, Table \ref{vicuna-13B}, Table \ref{LLaMA2-7B}, Table \ref{LLaMA2-13B}. 
The results indicate that our method performs well under different compression ratios and across various models.

At a fixed compression ratio, larger models exhibit less performance degradation after compression. Under low and high compression ratios, the performance of the compressed model exhibits markedly different characteristics. We first analyzed the performance of the model under low compression ratios. Under the compression ratio of 10\%, the compressed model even exhibite performance improvement in certain task (e.g. LLaMA-13B on BoolQ, PIQA, OBQA tasks).  This suggests that the appropriate low-ratio compression may improve the model's performance in certain tasks.  Under low compression ratios, the performance drop of the compressed model can be effectively recovered through fine-tuning.  However, under high compression ratios, significant performance decline is inevitable even after fine-tuning the model. Especially when the compression ratio reaches 40\%, the performance of the compressed model declines faster.

\section{Implementation Details}
\subsection{For Compression}
\textbf{Compression Ratio.} During the compression process, we only compress the transformer layers in the model , without applying any modifications to the embedding layers and the LM head. Therefore, to achieve the specified compression ratio for the model, we need to apply a higher compression ratio to the transformer layers. The formulation is: $$Ratio_{l}=\frac{Param_{total}\times Ratio_{s}}{layers\times Param_{layer}}$$
where $Param_{total}$ denotes the total number of parameters in the model and $Param_{layer}$ represents the compressible parameters within one transformer layer. $Ratio_{s}$ and $Ratio_{l}$ represent the specified compression ratios for the model and the actual compression ratio at the layer level, respectively. $layers$ denotes the number of transformer layers in the model. Taking the LLaMA model used in our experiments as example, we give the relationship between the specified model compression ratio $Ratio_{s}$ and the actual layer compression ratio $Ratio_{s}$ in the table  \ref{compression ratio}.
\vskip -0.2in
\begin{table}[htb]
\caption{The relationship between  $Ratio_{s}$ and $Ratio_{s}$ across LLaMA models of different sizes.}
  \centering
  \scriptsize
  \vskip 0.1in
  \begin{tabular}{c|ccccccccccc}
    \toprule
    $Ratio_{s}$ & 0.100  & 0.200 & 0.300 & 0.400 & 0.500 & 0.600 & 0.700 & 0.800 \\
    \midrule
    7B $Ratio_{l}$ & 0.104  & 0.208 & 0.312 & 0.416 & 0.520 & 0.624 & 0.728 & 0.832 \\
    13B $Ratio_{l}$ & 0.103  & 0.205 & 0.308& 0.410 & 0.513 & 0.616 & 0.718 & 0.821 \\
    30B $Ratio_{l}$ & 0.101  & 0.203 & 0.304 & 0.405 & 0.507 & 0.608 & 0.709 & 0.811 \\
    \bottomrule
  \end{tabular}
\vskip -0in
  \label{compression ratio}
\end{table}

\textbf{Compression Sequence.}
During the compression process, we independently compress the model layer by layer in sequence. The input activations are computed during a single forward computation. The output of the compressed layer serves as the input for the next layer in the model.

\textbf{Special Case.}
Due to the parameter allocation method, we tend to allocate more parameters to $\mathbf{W}_{v}$ and $\mathbf{W}_{o}$ matrices. Therefore, at low compression ratios , we may allocate more parameters to $\mathbf{W}_{v}$ and $\mathbf{W}_{o}$ matrices than the original weight matrix. In this case, we choose to directly retain the original weight matrix and allocate the surplus parameters to $\mathbf{W}_{q}$ and $\mathbf{W}_{k}$ matrices. 
For example, at the compression ratio of 20\%, the operation performed during MHA compression is to completely retain the $\mathbf{W}_{v}$ and $\mathbf{W}_{o}$ matrices, and substitute the $\mathbf{W}_{q}$ and $\mathbf{W}_{k}$ matrices with low-rank matrices which consist of 60\% of the parameter count.
\subsection{For Knowledge Recovery.}
We follow the fine-tuning method of LLM-pruner \cite{LLM-pruner} in the knowledge recovery phase. The hyperparameters are summarized in Table \ref{hyperparameters}. We use the hyperparameters in Table \ref{hyperparameters} to recovery the performance of the compressed model
\newpage
\begin{table}[htb]
\caption{The hyperparameters employed during the fine-tuning stage.}
  \centering
  \scriptsize
  \vskip 0.1in
  \begin{tabular}{ccccccccc}
    \toprule
    Optimizer & Epochs & Batch size  & Learning rate & Val-size & LoRA-r & LoRA-alpha &LoRA-dropout & LoRA-modules \\
    \midrule
   Adam & 2 & 64  & 1e-4 & 2000 & 8 & 16 & 0.05 & All  \\
    \bottomrule
  \end{tabular}
\vskip -0.1in
  \label{hyperparameters}
\end{table}
\section{Limitations}
During the compression process, we applied the same compression ratio to different transformer layers, which overlook the differences between layers in the model.
When the compression rate is high (exceeding 40\%), even after fine-tuning, the performance of the compressed model still exhibits a substantial decline. Achieving higher compression ratios for LLMs remains a challenging task. How to effectively restore the performance of the compressed model under limited resources is also a noteworthy concern.

\section{Generations From Compressed Model}
Table \ref{generate-llama}, Table \ref{generate-llama2}, Table\ref{generate-vicuna} show generated examples of the models compressed by LoRAP. We present the generation results of both the compressed model with fine-tuning and without fine-tuning in the same generation settings.
We can see that the compressed model at low compression ratio can generates fluent sentences, but the meaning of the generated sentences is different from the original model. However at high compression ratio the compressed model may generate repetitive or even semantically incorrect  sentences without fine-tuning.

\begin{table*}[htb]
\caption{Generated Examples from the Compressed LLaMA-7B.}
  \centering
  \scriptsize
  \vskip 0.1in
  \begin{tabular}{m{1.2cm}<{\centering}|m{14cm}}
    \toprule
    Model & Generated Sentences  \\
    \midrule
    Ratio=0\% wo/ Tune
   & \textbf{AI can assist people in writing} 100,000-word novels in just a few hours. AI is also being used to create art. In 2017, Google’s DeepDream AI was used to create a series of paintings that were eerily similar to the work of Dutch artist M.C. Escher. In 2018, an AI program created a painting that sold for \$432,500 at Christie’s auction house in New York City. The painting was created by an AI program called GAN, which stands for Generative Adversarial Networks. GAN is a type of AI that learns to mimic the style of another artist. In this case, the AI learned to mimic Vincent van Gogh’s style of painting. AI can also be used to create music. AI has been used \\ 

\midrule
 Ratio=0\% w/ Tune 
    & \textbf{AI can assist people in writing} 50,000-word research papers in a matter of hours. AI is also being used in the healthcare industry to assist doctors in diagnosing and treating patients. IBM’s Watson, for example, is being used in hospitals around the world to help doctors make more accurate diagnoses and prescribe more effective treatments. In the finance industry, AI is being used to analyze large amounts of data and identify patterns and trends that can help financial institutions make better investment decisions. AI is also being used to automate tasks such as trading stocks and managing portfolios, freeing up human workers to focus on more complex and strategic tasks. In marketing and advertising, AI is helping businesses to better understand their customers and to personalize their marketing efforts. \\

\midrule
   Ratio=20\% wo/ Tune
   & \textbf{AI can assist people in writing} 100,000 words in an hour. AI can also be used to improve the quality of products and services. For example, AI can be used to detect defects in manufacturing processes. AI can also be used in customer service. For example, if a customer calls a company’s customer service line, AI can analyze the customer’s voice to determine the customer’s mood. If the customer is in a bad mood, AI can route the customer’s call to a customer service representative who is trained to handle customers in a good mood. AI is also being used to develop new products. For example, companies are using AI to develop new products faster. For example, a company can use AI to develop a new product faster by using AI to design the product’s shape, size, and color. A company can also use AI to design a new \\ 
\midrule
 Ratio=20\% w/ Tune
    & \textbf{AI can assist people in writing}, editing, and proofreading their own work. Amazon’s Kindle Direct Publishing (KDP) is a self-publishing platform that allows authors to publish and sell their books directly to readers around the world. KDP provides free tools and services to help authors self-publish and earn royalties on their eBooks and print books sold through Amazon.com and Amazon.co.uk. Authors can also use KDP to create and publish customized versions of their books for specific audiences, such as students, employees, or members of a club or organization. For more information, visit www.kdp.amazon.com.
The National Association of College and University Business Officers (NACUB) is a national advocacy and professional development organization representing more than 1,500 colleges and universities.
 \\
\midrule
  Ratio=50\% wo/ Tune
   & \textbf{AI can assist people in writing} 100,000 articles a year. “We’ve been working on it for a couple of years, and now we’re getting closer and closer to making it a reality,” he said.\\ 
\midrule
    Ratio=50\% w/ Tune
     & \textbf{AI can assist people in writing} 50,000 words per hour, which is about 10 times faster than a human writing at their best. This means that in 24 hours, a human would have written 5000 words, whereas an AI could have written the same amount of text in just 1.5 hours. This is an incredible time-saving advantage for those who need to get things done quickly and accurately. You are here: Home  About Us  Our People Our people are the heart and soul of our business. They are our family, our friends, our colleagues, and our clients. We take great pride in them, and they take great pride in us. Our goal is to provide the highest level of service to our clients, and we strive to do so every day. We listen to our clients, understand their needs, and work hard to meet their expectations.\\
  \bottomrule
  \end{tabular}
\vskip -0.1in
  \label{generate-llama}
\end{table*}
\newpage
\begin{table*}[!htb]
\caption{Generated Examples from the Compressed LLaMA2-7B.}
  \centering
  \scriptsize
  \vskip 0.1in
  \begin{tabular}{m{1.2cm}<{\centering}|m{14cm}}
    \toprule
    Model & Generated Sentences  \\
    \midrule
    Ratio=0\% wo/ Tune
   & \textbf{People are currently exploring Mars} with the Curiosity rover, and NASA is planning to send astronauts to the Red Planet in the 2030s. NASA’s Mars 2020 rover will search for signs of past microbial life, characterize the planet’s climate and geology, collect samples for future return to Earth, and pave the way for human exploration of the Red Planet. NASA’s Perseverance rover will land on Mars on Feb. 18, 2021, to search for signs of ancient microbial life and collect rock and regolith (broken rock and dust) samples for possible return to Earth. The rover will characterize the planet's geology and past climate and be the first mission to collect and cache a sample of Mars to be returned to Earth in the future. Perseverance is part of NASA’s \\ 

\midrule
 Ratio=0\% w/ Tune 
    & \textbf{People are currently exploring Mars} with the Curiosity rover, and NASA is planning to send astronauts to the Red Planet in the 2030s. NASA’s Perseverance rover is currently exploring Jezero Crater on Mars, searching for signs of ancient microbial life and collecting samples of rock and regolith (broken rock and dust) that could be returned to Earth in the future. NASA’s Mars 2020 Perseverance mission is part of a larger program that includes missions to the Moon as a way to prepare for human exploration of the Red Planet. Challenges still remain, including how to keep astronauts healthy during long-duration spaceflight and how to keep them safe on the surface of another planet.\\

\midrule
   Ratio=20\% wo/ Tune
   & \textbf{People are currently exploring Mars}, the Moon, and other planets in our Solar System. They are also exploring the depths of our own oceans, and even the depths of their own minds. This is an exciting time to be alive. We are living in an age of discovery and exploration. We are discovering new worlds, and we are exploring our own minds. We are also living in an age where technology is advancing at an unprecedented rate. We have access to more information than ever before, and we have access to technology that would have been unimaginable just a few years ago. All of this means that we are living in a time of great opportunity. We have the opportunity to discover new worlds, to explore our own minds, and to take advantage of the technological advances that are happening all around us. So, what are you waiting for? Get out there and start.\\ 
\midrule
 Ratio=20\% w/ Tune
    & \textbf{People are currently exploring Mars} in unprecedented detail with the help of NASA's Mars rover Curiosity, which landed on the Red Planet in August 2012. The rover has been exploring the Martian surface for more than five years now, and it has sent back a wealth of information about the planet's geology, atmosphere, and habitability. One of the most exciting discoveries made by Curiosity was the detection of methane on Mars. Methane is an organic compound that can be produced by living organisms, and its presence on Mars has led scientists to speculate about the possibility of life on the planet. In a new study published in the journal Science, a team of researchers from NASA's Jet Propulsion Laboratory (JPL) and the California Institute of Technology (Caltech) has analyzed data from Curiosity's methane detector \\
\midrule
  Ratio=50\% wo/ Tune
   & \textbf{People are currently exploring Mars}. Mars are currently exploring Earth. People are currently exploring the Earth. The people are currently exploring on the Earth.\\ 
\midrule
    Ratio=50\% w/ Tune
    & \textbf{People are currently exploring Mars}, and they have discovered that the red planet is not as red as they thought it would be.
Mars has been explored since 1953 by the United States Air Force, NASA, and the European Space Agency (ESA). In 2007, NASA’s Spirit lander landed on Mars for the first time, and returned to Earth in 2011. Since then, more missions have been sent to Mars, including the Curiosity lander, which is still exploring the planet. In 2013, NASA’s Rover landed on Mars, becoming the first spacecraft to explore the Red Planet. The rover was able to gather samples of Martian soil, rocks, and dust, and send them back to Earth for further analysis. In July 2014, NASA’s Curiosity landed on Mars again, becoming the second\\
  \bottomrule
  \end{tabular}
\vskip -0.2in
  \label{generate-llama2}
\end{table*}

\begin{table*}[!htb]
\caption{Generated Examples from the Compressed Vicuna-7B.}
  \centering
  \scriptsize
  \vskip 0.1in
  \begin{tabular}{m{1.2cm}<{\centering}|m{14cm}}
    \toprule
    Model & Generated Sentences  \\
    \midrule
    Ratio=0\% wo/ Tune
   & \textbf{Last week, we went skiing together} for the first time in a long time, and I was reminded of how much I love the feeling of gliding down freshly powdered slopes, the crisp winter air filling my lungs, and the sense of freedom and exhilaration that comes with hurtling down a mountain at breakneck speeds. hopefully, we'll be able to go skiing together again soon, and I'll be able to experience that rush of adrenaline all over again. In the meantime, I'll just have to remember the feeling and hold onto it tightly, like a precious memory that I can revisit whenever I need a little pick-me-up. \\ 

\midrule
 Ratio=0\% w/ Tune 
    & \textbf{Last week, we went skiing together} for the first time in a long time, and it was so much fun! The freshly powdered slopes were crowded with skilled skiers gliding down the freshly powdered slopes, and the crisp winter air was invigorating as we laughed and chatted our way down the freshly powdered slopes. hopefully, we'll be able to do it again soon.\\

\midrule
   Ratio=20\% wo/ Tune
   & \textbf{Last week, we went skiing together} for the first time in over a year. hopefully, we'll be able to do it more often now that we're both vaccinated. It was great to be able to hit the slopes again after such a long hiatus, and I'm looking forward to many more skiing trips in the future. What about you? Have you been able to get back to any of your favorite activities yet, or are you still waiting for your second dose of the vaccine?\\ 
\midrule
 Ratio=20\% w/ Tune
    & \textbf{Last week, we went skiing together} for the first time, and it was amazing! The snow was perfect, the slopes were crowded but not too crowded, and the scenery was breathtaking. hopefully, we'll be able to do it again soon. What about you? Have you ever gone skiing or snowboarding? If not, would you like to try it out? Let me know in the comments below!  \\
\midrule
  Ratio=50\% wo/ Tune
   & \textbf{Last week, we went skiing together} for the first time, and I have to say, it was an amazing experience! We had a great deal of fun on the slopes, turns, and curves, and it was a lot of fun! We had a great deal of fun on the slopes, turns, and curves, and it was a lot of fun! We had a great deal of fun on the slopes, turns, and curves, and it was a lot of fun! We had a great deal of fun on the slopes, turns, and  \\ 
\midrule
    Ratio=50\% w/ Tune
    & \textbf{Last week, we went skiing together} for the first time and it was amazing! We had so much fun on the freshly powdered slopes and traveled down the freshly powdered slopes with our skis. The next day, we went hiking in the mountains and it was beautiful! We hiked up to the top of the mountain and had a breathtaking view of the valley below. The third day, we went to the beach and it was perfect! We swam in the cool water and\\
  \bottomrule
  \end{tabular}
\vskip -0.1in
  \label{generate-vicuna}
\end{table*}


\end{document}